\pdfoutput=1

\documentclass[11pt]{article}

\usepackage{ACL2023}

\usepackage{times}
\usepackage{latexsym}
\usepackage{pifont}
\usepackage{newunicodechar}
\usepackage{algorithm}
\usepackage{multirow}
\usepackage{array}
\usepackage{subcaption}
\usepackage{bbm}
\usepackage{enumitem}
\usepackage{booktabs}     
\usepackage[noend]{algpseudocode}

\usepackage[T1]{fontenc}

\usepackage[utf8]{inputenc}

\usepackage{microtype}
\usepackage{amsmath}
\usepackage{amssymb}
\usepackage{graphicx}
\usepackage{xspace}

\usepackage{inconsolata}
\newunicodechar{✓}{\ding{51}}
\newunicodechar{✗}{\ding{55}}

\newcommand{\ours}{BalDistill\xspace}

\usepackage{ulem}

\title{Multi-Stage Balanced Distillation: Addressing Long-Tail Challenges in Sequence-Level Knowledge Distillation}

\author{Yuhang Zhou$^{\ 1*}$\quad Jing Zhu$^{\ 2*}$ \quad Paiheng Xu$^{\ 1}$ \quad Xiaoyu Liu$^{\ 1}$ \quad Xiyao Wang$^{\ 1}$\quad \\ \textbf{Danai Koutra$^{\ 2}$}\quad \textbf{Wei Ai$^{\ 1}$} \quad \textbf{Furong Huang$^{\ 1}$}\\
         $^{1}$ University of Maryland, College Park \\ $^{2}$ University of Michigan, Ann Arbor \\ \texttt{\{tonyzhou, paiheng, xliu1231, xywang, aiwei, furongh\}@umd.edu} \\
         \texttt{\{jingzhuu, dkoutra\}@umich.edu}}

\begin{document}
\maketitle
\begin{abstract}


Large language models (LLMs) have significantly advanced various natural language processing tasks, but deploying them remains computationally expensive. Knowledge distillation (KD) is a promising solution, enabling the transfer of capabilities from larger teacher LLMs to more compact student models. Particularly, sequence-level KD, which distills rationale-based reasoning processes instead of merely final outcomes, shows great potential in enhancing students' reasoning capabilities. However, current methods struggle with sequence-level KD under long-tailed data distributions, adversely affecting generalization on sparsely represented domains. We introduce the Multi-Stage Balanced Distillation (\ours) framework, which iteratively balances training data within a fixed computational budget. By dynamically selecting representative head domain examples and synthesizing tail domain examples, \ours achieves state-of-the-art performance across diverse long-tailed datasets, enhancing both the efficiency and efficacy of the distilled models.  \footnote{Our code and data are available at \url{https://github.com/Tonyzhou98/long_tail_kd}}

\let\thefootnote\relax\footnote{*Equal contribution.}
\setcounter{footnote}{0}

\end{abstract}

\section{Introduction}

Large language models (LLMs) like GPT-4 and LLaMA have revolutionized tasks ranging from text generation to language translation through their deep understanding and generation of human-like text~\cite{openai2023gpt4, touvron2023llama, vicuna2023, jiang2023mistral}. Despite their success, the deployment of these models is hindered by their substantial size and computational demands, especially in environments with limited resources. Knowledge distillation (KD) offers a viable solution by transferring knowledge from expensive teacher models to smaller, efficient student models. Specifically, \textit{sequence-level KD} focuses on distilling rationale-based reasoning processes rather than final outcomes.
It leverages the teacher's reasoning processes, encapsulated in chain-of-thought (CoT) rationales, to enhance the student models’ generative capabilities~\cite{kim2016sequence, ho2022large, shridhar2022distilling, hsieh2023distilling}.

However, there are a few challenges to fully leverage the power of sequence-level KD, as follows.
\textbf{(C1)} Sequence-level KD encounters significant challenges when training with long-tailed data distributions, which are prevalent in real-world scenarios --- data often follows a power-law distribution with a few dominant classes (head) and many rare classes (tail) ~\cite{liu2019large}. Such distributions feature a few dominant classes and many underrepresented ones, leading to models that generalize poorly on sparsely represented domains. 
\textbf{(C2)} Traditional KD methods in the text area to solve long-tail challenges, often reliant on direct access to model weights or loss adjustment primarily suited for straightforward classification tasks \cite{zhou2023scalable, schick2021s, dai2023long, zhang2022making, tepper2020balancing}, falter under the complexities of sequence-level KD, especially when the teacher model is a black box and the task is generative, which is our target. 
\textbf{(C3)} Addressing this imbalance is critical, yet resource-intensive, as it typically requires generating a large volume of synthetic data to balance the dataset~\citet{tepper2020balancing}. Moreover, naively up-sampling the long-tailed dataset may dramatically increase the number of calls to the teacher models. Budget constraints play a crucial role in KD for black-box LLMs, as querying the teacher for rationales can be costly and time-consuming ~\cite{chen2023frugalgpt, zhou2024teachingassistantintheloop}.

\begin{figure*}[!ht]
  \centering
  \includegraphics[width=0.9\textwidth]{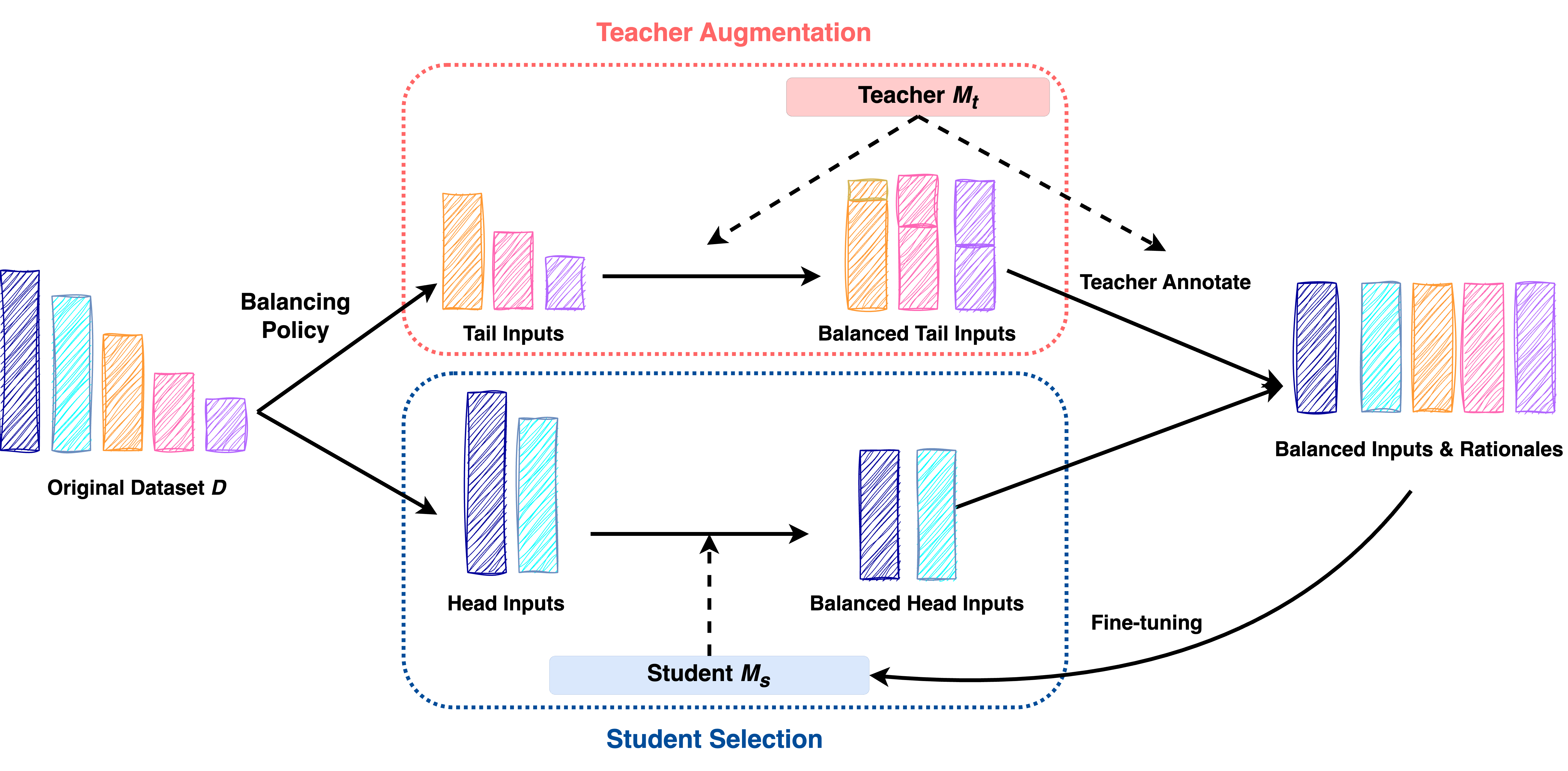}
  \vspace{-1em}
   \caption{\label{fig:overview} \textbf{Overview of the proposed iterative \ours framework.} The framework is composed of multiple stages. For each stage, we apply the balancing policy to decide the data distribution in the training batch. For head domains with sufficient data, we actively extract the examples by IFD metrics using the student model. For the tail domains, we call the teacher model to generate the synthetic examples and the corresponding rationales. The teacher model finally annotates the balanced training batch and fine-tunes the student model. 
   }
   \vspace{-0.3cm}
\end{figure*}

Our proposed solution, the Multi-Stage Balanced Distillation (\ours), tackles all the challenges above by strategically generating balanced training sets within budget constraints and iteratively fine-tuning the student model with actively selected and synthetic data for multiple stages.
\ours progressively refines the training data by selecting key examples from well-represented domains and generating necessary synthetic data for underrepresented ones, ensuring comprehensive domain coverage and model robustness.
By dynamically selecting representative head domain examples and synthesizing tail domain examples, \ours achieves state-of-the-art (SoTA) performance on various long-tailed datasets, enhancing both the efficiency and efficacy of the method.

Our contributions are summarized as follows: 
\begin{itemize}[leftmargin=*]
    \item \textbf{Innovative Problem Framing:} We address the under-explored challenge of applying sequence-level KD to long-tailed distributions, where the teacher model is a black-box LLM.

    \item \textbf{Strategic Framework:} \ours innovatively combines active example selection with synthetic data generation for multiple stages to maintain training balance within predefined budget limits.
    
    \item \textbf{SoTA Performance:} Our framework demonstrably improves the student models' effectiveness and robustness across diverse domains, setting new benchmarks in performance. We empirically demonstrate that our distilled student models achieve state-of-the-art performance across a range of benchmark datasets.

\end{itemize}

\section{Related Work}

\noindent \textbf{Knowledge Distillation} uses the outputs of a larger LLM (Teacher), such as ChatGPT \cite{openai2023gpt4}, to train a smaller model (Student), such as LLaMa-7B \cite{touvron2023llama}. For details of knowledge distillation (KD) of large language models, we refer to the survey for more details ~\cite{xu2024survey}. In this work, we focus on KD with black-box teacher models. There are two lines of work with respect to knowledge distillation. The first is to ask teacher models to generate the final answers and to fine-tune on the final answers ~\cite{zhou2023scalable, schick2021s}. Another line of work asks teacher models to generate rationales at the reasoning process and fine-tunes student models on the rationales in the sequence level to improve their reasoning ability \cite{ho2022large, shridhar2022distilling, hsieh2023distilling}, which proves to be more effective. In this work, we mainly discuss using a teacher model to generate rationales and improve the student's reasoning ability on a long-tailed dataset. Despite the progress of KD in the LLM era, existing works fail to establish a pipeline to gain knowledge from long-tailed datasets with the sequence-level KD, as few rationale examples are provided for tail knowledge.

\vspace{0.5em}
\noindent \textbf{Long-Tail Learning} focuses on long-tail
distributed data and has been an emerging topic of interest in the NLP community ~\cite{liu2019large, wang2017learning, godbole2022benchmarking, dai2023long, zhang2022making, liu2024large, mondal2024scidoc2diagrammermafgenerationscientificdiagrams}. Approaches to solving the long-tail problem include rebalancing, information augmentation, and module improvement ~\cite{zhang2021videolt, he2021distilling, zhu2023pitfalls, cui2021parametric, xu2024promises}. Despite the importance of long-tail learning, studies have shown that LLMs struggle to learn long-tail knowledge ~\cite{kandpal2023large,sun2023head}. In this work, we propose to improve LLMs' ability to learn long-tail knowledge via multi-stage distillation over balanced datasets. 

\vspace{0.5em}
\noindent \textbf{Active Learning} aims to reduce labeling effort by selecting only the most useful examples. Traditional active learning can be categorized into uncertainty-based methods ~\cite{prabhu2019sampling, margatina2021active, wang2022live} and diversity-based methods ~\cite{ru2020active, ash2019deep}. In the LLM era, active learning has been used to reduce human annotation costs by (1) strategically selecting the most informative examples for human feedback or annotation ~\cite{margatina2023active, osband2022fine, wang2020neural} and (2) integrating language models as annotators within an active learning framework without human supervision ~\cite{xiao2023freeal, rouzegar2024enhancing, zhang2023llmaaa, li2024improving,liu2024csrec}. In this work, we propose to solve the long-tail problem in the student LLM by actively distilling knowledge from a black-box teacher LLM to meet the budget requirement.  
\section{Methodology}

\label{sec:method}

\subsection{Problem Statement}

We define our research problem as follows: Given the teacher LLM ($\mathcal{M}_t$), the student LLM ($\mathcal{M}_s$), a long-tailed dataset $\mathcal{D}$ (with domain number $[d_1, d_2, \dots, d_l]$ for $l$ domains in total) and a fixed budget $B$ to query the teacher, we seek to propose an efficient framework to fine-tune an effective and robust student model, $\mathcal{M}_s$, over $\mathcal{D}$.

\subsection{Overall Approach}

To mitigate the performance bias in KD caused by long-tailed datasets within budget constraints, we employ a strategy that combines synthetic data augmentation with active selection. This approach ensures effective fine-tuning across both well-represented (`head') and underrepresented (`tail') domains.
As depicted in Figure \ref{fig:overview}, we propose a \textit{multi-stage} framework to create the training data \textit{iteratively}. 

We operate in a pool-based setting where a large dataset, denoted as $\mathcal{D}$, is available but lacks annotations from a teacher model.

At each stage of our \ours process, we first implement a balancing policy, which we have designed, to determine the appropriate data distribution for each domain within the training batch. This policy is based on the principles of data equality and training effectiveness across domains, aiming to optimize learning outcomes despite data scarcity in certain areas. The total number of stages is predefined based on the consideration of efficiency and the optimal performance.

For domains well-represented in our dataset $\mathcal{D}$ (referred to as `head domains'), we employ active selection techniques \cite{touvron2023llama, yuan2020cold} using the fine-tuned student model $\mathcal{M}_s$ to identify and extract the most informative examples from the pool. 
Conversely, for domains lacking sufficient data (`tail domains'), we utilize the teacher model $\mathcal{M}_t$ to generate both synthetic samples and corresponding annotations, enriching the training material available.

After selecting and/or generating these samples, we query the teacher model to provide detailed rationales for examples in the training batch. These annotated examples are then used to fine-tune the student model $\mathcal{M}_s$ in preparation for the next stage.  
Detailed descriptions of these components, along with the algorithms outlining this procedure, are presented in Algorithm~\ref{alg:method}.

\begin{algorithm}[t!]
\caption{\textbf{Multi Stage Balanced Distillations}}
\label{alg:method}
\begin{algorithmic}[1]

\State \textbf{Input:} Long tailed dataset $D$, Student model $M_{s}$, Teacher model $M_{t}$, prompt for generating data $P_c$, Stage number $K$, Balancing policy $P$, Training bucket $T$, Budget number $B$
\State \textbf{Output:} The fine-tuned student model $M_{s}^K$

\For{each stage $k = 0, \dots, k-1$}
\State $\text{head, tail domains} = P(D, k, B)$

\For{each domain tail domain $j$}
\State Add remaining $x_{j}$ from $D$ to $T$
\State $\hat{x_{j}} = M_t(P_c, j)$
\State Add synthetic $\hat{x_{j}}$ to $T$
\EndFor

\For{each domain head domain $h$}
\State Collect all $x_h$ from $D$
\State $x_h = M_{s}^{k-1} (x_h, h)$
\State Add selected $x_{h}$ to $T$
\EndFor

\State Use $M_t$ to annotate $x$ in $T$ w/o rationales
\State $M_s^k = \text{Fine-tune}(M_s, T)$
\EndFor
\end{algorithmic}
\end{algorithm}

\subsection{Balancing Policy}

Considering $K$ total stages in our framework, we first evenly divide our budget $B$ into $K$ parts, which means that for each stage, we create a small training batch with $\frac{B}{K}$ examples extracted from $\mathcal{D}$ with teacher-annotated rationales.
Within a small training batch, we propose two strategies to allocate the budget over different domains. 

\paragraph{Naive Balancing} Since our goal is to mitigate the bias towards head domains, our first balancing policy is to use naive balancing, which selects the same number of inputs for each domain in the training batch. Formally, the number of samples for each domain in the small training batch is $\frac{B}{Kl}$, where $l$ is the number of domains in the dataset.

\paragraph{Adaptive Balancing} 

One of our staged learning framework's key features is utilizing the fine-tuned student model to actively select representative inputs from well-represented domains, known as head domains. However, employing a naive balancing policy typically results in the disproportionate allocation of the training budget to data from underrepresented domains, or tail domains. 
This training batch may lead the fine-tuned student model to struggle to select truly effective examples from the head domains, particularly in the initial stages. Such selections are crucial for the model to learn effectively from these domains. To address this, we implement an adaptive balancing policy. This policy starts by constructing the training batch with a distribution akin to random selection, thus primarily focusing on head data in the early stages to `warm up' the model. As the process advances, the policy gradually shifts towards a more balanced distribution by the final stage, ensuring comprehensive learning across both head and tail domains.

Formally, the number of examples for each domain is the weighted average between the numbers for random selection and the numbers for naive balancing. For stage $i$, domain $d$, we select $(\frac{n_d}{N} \cdot \frac{B}{K}) \cdot \frac{K - i}{K} + \frac{B}{Kl} \cdot \frac{i}{K}$ examples for domain $d$ to build the training batch for adaptive balancing, where $N$ and $n_d$ are the total number and the domain size in the original data $\mathcal{D}$. 

Then, domains are naturally categorized based on whether the number of required samples per domain exceeds the available samples in the pool. Domains requiring more samples than available are designated as `head domains' for that particular stage, while those with fewer required examples than available are categorized as `tail domains.'

For tail domains, where there are insufficient samples in the dataset $\mathcal{D}$, we rely on the teacher model to generate both the samples and their corresponding rationales, detailed in Section~\ref{sec:teacher_aug}. 
In contrast, for head domains, which have a sufficient number of samples available to meet the demands of the training batch, we utilize the fine-tuned student model to actively select the most representative samples, as discussed in Section~\ref{sec:student_active}.

It is important to note that the classification of domains as head or tail can vary across different stages of the training process, depending on the evolving needs and data availability.

\subsection{Teacher Data Augmentation}
\label{sec:teacher_aug}
Motivated by the effectiveness of synthetic dataset generated by black-box LLMs \cite{openai2023gpt4, radford2019language, zhou2024explore}, we utilize the teacher LLMs to generate synthetic samples and corresponding annotations to upsample data for tail domains. To save the annotation budget, we require the teacher model to compose the sample and the corresponding rationales at the same time. 

Suppose that we need $m$ synthetic examples for domain $a$ to satisfy the training batch requirement. Given an instruction following prompt $P_c$, composed of three demonstrations from domain $a$, and teacher model $\mathcal{M}_t$, we employ stochastic temperature sampling with a fixed temperature and repeat the process $m$ times with generated samples $\hat{x}_{a1}, \cdots \hat{x}_{am}$ and rationales $\hat{y}_{a1}, \cdots \hat{y}_{am}$:
\begin{equation*}
    \hat{x}_{ai}, \hat{y}_{ai} =M_{t}(P_c, a) \quad\text{for $i\in\{1,\cdots, m\}$ }
\end{equation*}

Then we add the generated samples and rationales to the training batch and combine with the extracted samples from $\mathcal{D}$. We present two examples of synthetic inputs and rationales from the teacher model in Table \ref{table:synthetic_examples} in Appendix \ref{sec:implementation}. The case study suggests the effectiveness of the teacher model in generating tail examples.

\subsection{Student Active Selection}
\label{sec:student_active}

For head domains, our strategy involves actively selecting instances from the original dataset to meet the numeric requirements of the balancing policy. We aim to mitigate information loss from data downsampling through this active data acquisition. The objective is to identify the most challenging or uncertain instances for the student model, thereby optimizing its learning trajectory.

To quantify instance uncertainty, we adapt the Instruction Following Difficulty (IFD) metric originally proposed by \citet{li2024superfiltering, li2024quantity}. The IFD scores are used to measure a training instance's uncertainty level as perceived by the student model. IFD is calculated as the ratio of the perplexity of generating a response $y$ with an input $x$ to the perplexity of generating $y$ without $x$:
$\text{IFD}(x, y) = \frac{\text{PPL}(y | x)}{\text{PPL}(y)}$,
where PPL represents perplexity, a metric widely used to evaluate language model performance \cite{jelinek1977perplexity}. 
Studies have shown that IFD scores offer greater efficiency in data selection compared to methods like K-means diversity or sole reliance on perplexity \cite{li2024superfiltering, settles2009active, yuan2020cold}.

A higher IFD score indicates an increased difficulty for the model in generating the response, highlighting the instance's value for training~\cite{li2024superfiltering}. 

Unlike the approach in~\citet{li2024superfiltering}, which utilizes ground-truth or advanced LLM-generated responses $y$, our setting imposes budge constraints that prevent such usage.
Instead, we calculate IFD using rationals $\hat{y}_s$ generated by the previously fine-tuned student model, allowing us to assess the model's self-uncertainty and conserve the annotation budget from the teacher model.

At last, we rank the inputs by their IFD scores, selecting those with the highest values to include in the batch, as specified by the balancing policy.

\subsection{Reasoning Generation and Fine-tuning}

Building on methodologies from prior research that focus on distilling reasoning abilities from black-box LLMs \cite{ho2022large, hsieh2023distilling}, we employ a zero-shot CoT approach, where the teacher model is prompted to generate a reasoning explanation $\hat{y}_t$ for the samples in our constructed training batch. This zero-shot setting is crucial for demonstrating the model's ability to reason based on its pre-existing knowledge alone~\cite{brown2020language}.
In our experimental setup, which utilizes labeled datasets lacking rationale annotations, the final ground truth answer is included in the prompt. 
This inclusion ensures that the generated explanations are aligned with the correct outcomes, enhancing the accuracy and relevance of the CoT reasoning. 
It is important to note that for synthetic samples generated from tail domains in~\ref{sec:teacher_aug}, we do not perform additional annotations in this part to maintain adherence to budget constraints.

After gathering the required samples and their associated rationales in the training batch, we integrate this batch with the annotated data accumulated from previous stages. This approach ensures that our student model is exposed to a diverse and comprehensive dataset, which helps mitigate the risk of overfitting --- a common challenge in machine learning models as identified in prior studies \cite{dor2020active,liu2023cat}. To facilitate this, we reinitialize and fine-tune the student model on the compiled rationale sequences from scratch at each stage.

The fine-tuning is performed using autoregressive language modeling with a cross-entropy loss, aligning with the original pre-training objectives of the student model \cite{touvron2023llama}. 

\section{Experiment}
\label{sec:experiments}

Through our extensive empirical analysis, we aim to address the following research questions:
\begin{itemize}[leftmargin=*]
    \item \textbf{RQ1}: How effective is our KD framework compared to previous KD baseline methods?
    \item \textbf{RQ2}: How important is each component (balancing policy and active learning) to the framework? 
    \item \textbf{RQ3}: How well does our method perform with different student models and budget restrictions? 
\end{itemize}

\paragraph{Dataset}
To verify the effectiveness of our framework on various reasoning tasks, we evaluate our method on five long-tailed datasets, following previous work \cite{yu2023metamath, dai2023long, huang2021balancing}. These include text classification: R52 and Reuters \cite{hayes1990construe}, question answering: AbstractiveQA and Multiple-choiceQA \cite{dai2023long} and arithmetic: MATH \cite{hendrycks2021measuring}. For text classification datasets, we treat the label of inputs as the domain; for other datasets, the domain information of inputs is annotated as metadata from the data provider. The detailed construction process and domain information for these datasets can be found in Appendix \ref{data_details}. We also show two example distributions of the datasets in Figure~\ref{fig:dataset-long} in Appendix \ref{data_details}. For each dataset, we prepare two budget settings for the experiment. In Table \ref{table:dataset}, we present the budget number, the test number, the domain number, and the evaluation metric of all five datasets.
The budget number in Table \ref{table:dataset} represents the total number of queries to the teacher models. For example, budget setting 1 for the R52 dataset is 2,600, which means that for our \ours method, the sum of queries to the teacher model for data augmentation and for generating reasoning steps should also equal 2,600 without incurring additional costs. This budget ensures that all operations, including data augmentation and reasoning step generation, are performed within the allocated query limit.

\paragraph{Evaluation metrics}

Since we are dealing with long-tailed imbalanced data, for each dataset, we choose to use both the micro- and macro-averages to evaluate the method robustness \cite{henning2022survey, li2024pedantspreciseevaluationsdiverse}. For the classification datasets (R52 and Reuters), we report micro-F1 and macro-F1, where micro-F1 is a global average F1 score and macro-F1 is computed by taking the unweighted mean of all the per-class F1 scores \cite{harbecke2022only}. For other datasets, we also report the micro-/macro-F1 for AbstractiveQA datasets and micro-/macro-accuracy for Math and Multi-choiceQA datasets. Note that the F1 score for the AbstractQA is the word-level F1 score between the token list of ground truth answer and the generated answer, different from the F1 for the classification task.

\paragraph{Model setup}

For the teacher model, we use GPT-4 \cite{openai2023gpt4} to generate the CoT rationales for each dataset. We choose between Llama2-7B and Llama3-8B as our student models \cite{touvron2023llama}.
We include the detailed configurations and implementations of the model in Appendix \ref{sec:implementation}.

\begin{table}[!t]
\centering
\resizebox{\columnwidth}{!}{%
\begin{tabular}{lccccc}
\toprule
Dataset         & \# Budget    & \# Test & \# Domain & Metric   & Task       \\ \midrule
R52             & 2,600/5,200  &  2,570       & 52        & F1       & TC         \\ 
Reuters         & 4,500/9,000  &  3,745       & 90        & F1       & TC         \\ 
Abstractive QA  & 5,000/10,000 &  10,000       & 5         & F1       & QA         \\ 
Multi-choice QA & 5,000/8,000  &  10,520       & 10        & Accuracy & QA         \\ 
Math            & 2,100/3,500  &  5,000       & 7         & Accuracy & Arithmetic \\ \bottomrule
\end{tabular}
}
\vspace{-1em}
\caption{\label{table:dataset} Dataset statistics. TC and QA represent the text classification and question answering, respectively.}
\vspace{-1em}
\end{table}

\paragraph{Baseline methods}
We experiment with two variants of our proposed method with different balancing policies, as discussed in Section \ref{sec:method}: In our first framework \textbf{\ours (N)}, we use naive balancing policy, and for second framework \textbf{\ours (A)}, we leverage adaptive balancing. We compare our framework with multiple baseline methods: (1) \textbf{Zero-shot CoT}. We directly prompt the student model to infer on the test data \cite{kojima2022large}. (2) \textbf{Random Finetune}. We randomly collect samples from the training data until the budget constraint is met and finetune student models on the final ground-truth labels \cite{radford2019language}. (3) \textbf{Random Finetune-CoT}. We randomly collect and use CoT rationales from the teacher model for student fine-tuning \cite{ho2022large, yao2022react, he2023teacherlm}. (4) \textbf{Duplicate Finetune-CoT}. We construct the training data with a naive balancing policy. For the tail domains, we duplicate the inputs to satisfy the policy requirement and for head domains, we randomly sample examples in over-represented domains. 
\begin{table*}[!htb]
\centering
\resizebox{\linewidth}{!}{%
\begin{tabular}{lcccccccccc}
\toprule
\multicolumn{1}{c}{\multirow{2}{*}{Method}} & \multicolumn{2}{c}{R52}         & \multicolumn{2}{c}{Reuters}     & \multicolumn{2}{c}{AbstractiveQA} & \multicolumn{2}{c}{Multi-choiceQA} & \multicolumn{2}{c}{Math}        \\                       & macro-f1       & micro-f1       & macro-f1       & micro-f1       & macro-f1        & micro-f1        & macro-acc        & micro-acc       & macro-acc      & micro-acc      \\ \midrule
Zero-shot CoT                               & 0.89           & 2.30           & 0.74           & 1.61           & 7.60            & 7.59            & 24.67            & 24.95           & 7.57           & 8.68           \\
Random Finetune                             & 45.95          & \textbf{91.44} & 28.01          & \textbf{74.68} & 37.62           & 37.21           & 61.23            & 55.96           & 10.12          & 9.48           \\
Random Finetune CoT                         & \textbf{59.70} & 89.46          & 27.35          & 70.53          & 52.57           & 52.88           & 76.09            & 74.12           & 16.62          & 15.20          \\
Duplicate Finetune CoT                      & 46.56          & 71.79          & 26.76          & 62.84          & 51.32           & 51.37           & 75.92            & 73.99           & 16.98          & 15.05          \\ \midrule
\ours (N)                              & 59.62          & 82.49          & 28.09          & 62.40          & 52.70           & \textbf{52.92}  & 76.60            & 73.43           & 17.90          & 16.34          \\
\ours (A)                              & 58.93          & 87.47          & \textbf{32.95} & 69.77          & \textbf{53.20}  & 52.90           & \textbf{77.17}   & \textbf{74.73}  & \textbf{18.66} & \textbf{17.42} \\ \bottomrule
\end{tabular}
}
\vspace{-0.5em}
\caption{\label{table:results} \textbf{Performance of proposed \ours framework and other baselines across five long-tailed datasets.} The best performance is marked in bold. The performance of fine-tuned student models with our framework can outperform other baselines in macro-averages on multiple long-tailed datasets.}
\end{table*}

\section{Results}

\subsection{Comparison with Baseline Methods}
\label{sec:overall_results}

\noindent \textbf{\ours framework outperforms Random Finetune and Duplicate Finetune methods.} \quad
We use Llama3 as the student model, GPT-4 as the teacher model, and choose the smaller budget for each dataset in Table \ref{table:dataset} as our experiment settings for this subsection. We present the overall macro- and micro-average results of the proposed frameworks and the baseline methods in Table \ref{table:results}. From Table \ref{table:results}, we first observe that on the long-tailed dataset, the methods fine-tuned on teacher-generated rationales (CoT) can significantly outperform the ground-truth fine-tuning method (Random Finetune), which emphasizes the necessity of teacher-generated reasoning steps in the KD.

Among all sequence-level KD methods, our proposed \ours (N) and \ours (A) achieve the best average performance across various datasets on macro-averages, which obtain an average relative improvement of 2.24\% and 6.81\%, respectively, compared to the Random Finetune CoT baseline. The performance boost in \ours (N) implies the effectiveness of replacing the naive balancing policy with adaptive balancing.

Moreover, we note that the Duplicate Finetune CoT baseline fails to compete with the Random Finetune CoT method in most cases, which indicates that simply duplicating the input from the tail domains to ensure balanced data cannot address the underlying imbalanced data complexity.

\begin{figure}[t!]
    \centering
    \begin{subfigure}[b]{0.48\columnwidth}
        \centering
        \includegraphics[width=\textwidth]{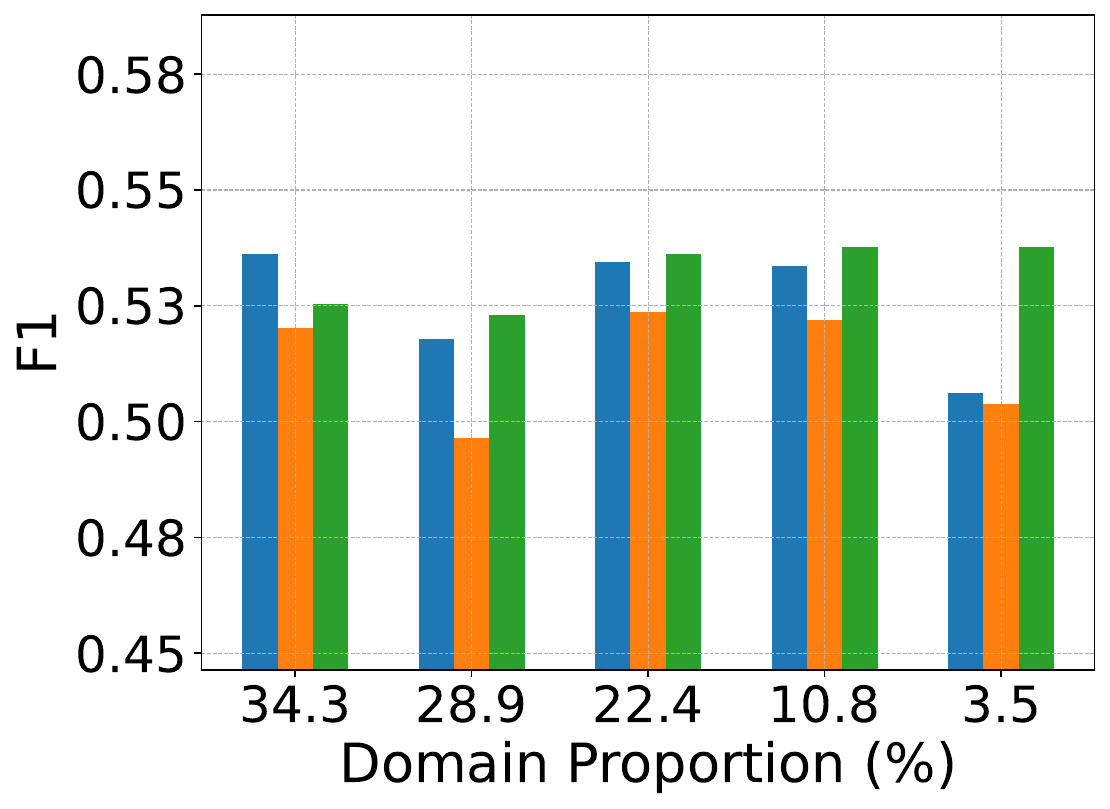}
        \caption{AbstractiveQA}
        \label{fig:abstractive_com}
    \end{subfigure}
    \hfill
    \begin{subfigure}[b]{0.48\columnwidth}
        \centering
        \includegraphics[width=\textwidth]{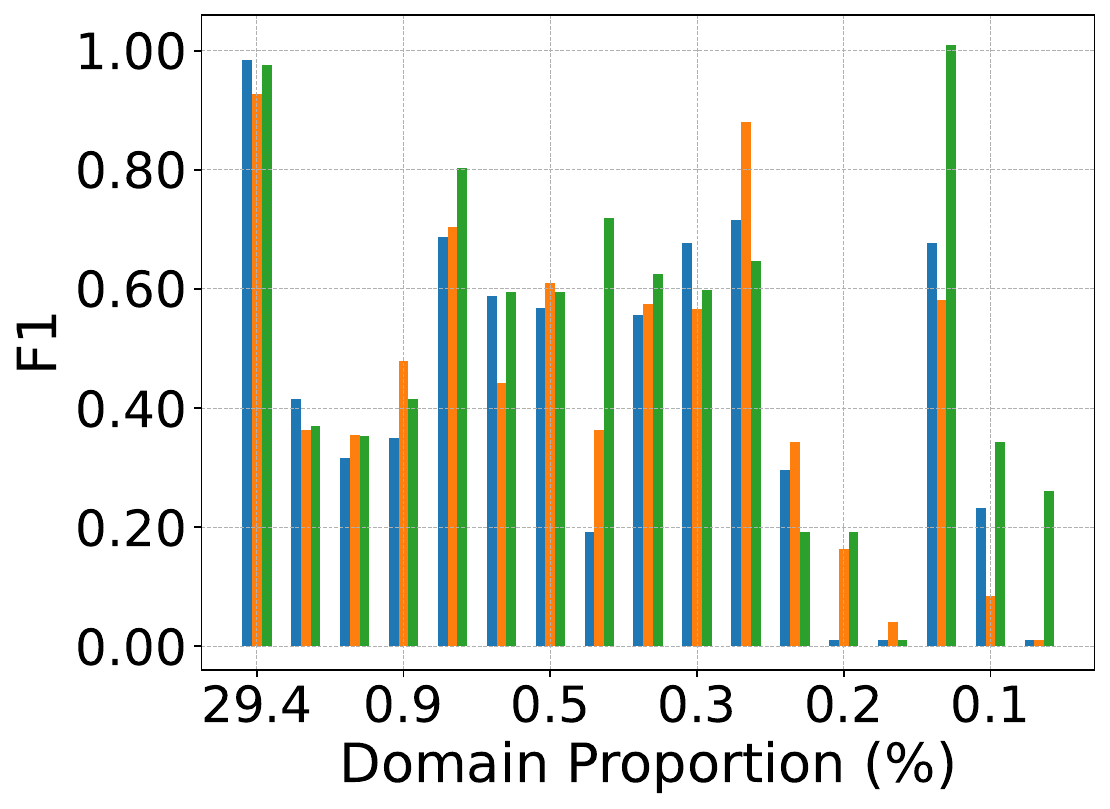}
        \caption{Reuters}
        \label{fig:reuters_com}
    \end{subfigure}
    \vfill
    \begin{subfigure}[b]{0.48\columnwidth}
        \centering
        \includegraphics[width=\textwidth]{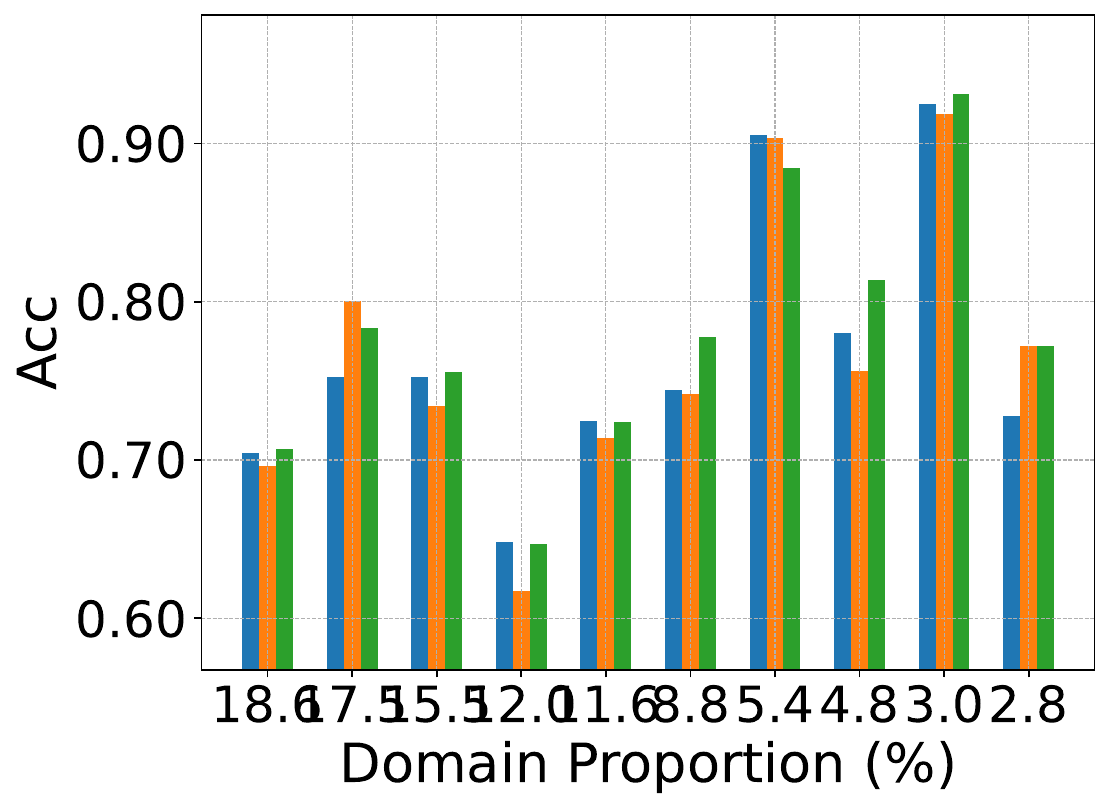}
        \caption{MultichoiceQA}
        \label{fig:multichoiceQA_com}
    \end{subfigure}
    \hfill
    \begin{subfigure}[b]{0.48\columnwidth}
        \centering
        \includegraphics[width=\textwidth]{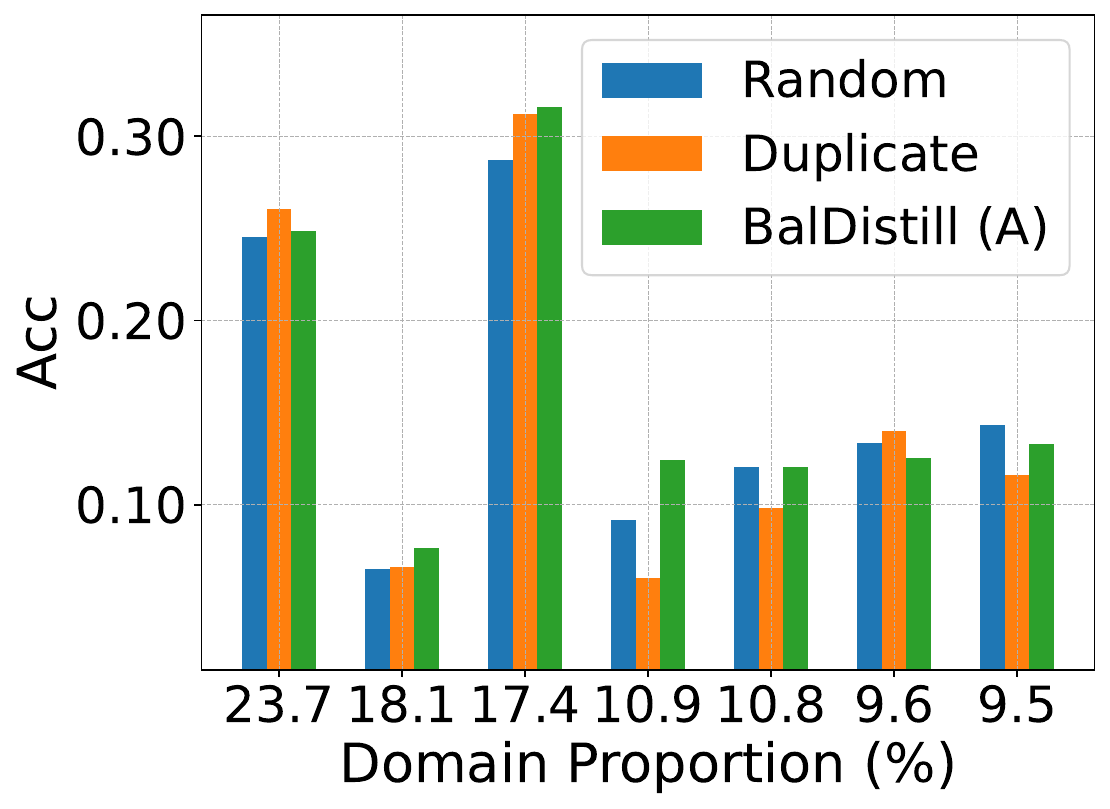}
        \caption{Math}
        \label{fig:math_com}
    \end{subfigure}
    \vspace{-0.6em}
    \caption{\textbf{Performance of proposed method and baselines on different domains.} X-axis represents the proportion of each domain, ranked from head to tail domains. Our proposed \ours method can achieve comparable results on head domains and outperform the baseline method on the tail domains. 
    }
    \label{fig:accuracy_comparison}
\end{figure}

To perform a detailed analysis of our framework, we visualize the F1 or accuracy score for each domain of the \ours (N) method and two baseline methods (Random Finetune CoT and Duplicate Finetune CoT) in Figure \ref{fig:accuracy_comparison}, with the x-axis representing the proportion of each domain in the dataset in descending order. From Figure \ref{fig:accuracy_comparison}, our proposed method can achieve comparable results in the head domains (left side of the figure) but substantially outperform the baseline methods in the tail domains (right side of the figure). This observation verifies our expectation in Section \ref{sec:method}, where the balancing policy increases performance in the tail domain, and the active learning part improves the data efficiency to compensate for data loss in the head domain. Note that for Math dataset, \ours can only achieve comparable results with the baseline methods on the last two tail domains (precalculus and probability), and we conjecture that the high difficulty in these two domains prevents the teacher from composing high-quality synthetic data. We also include 2 additional SoTA methods for multitask learning or resolving class imbalance challenges and compare their results with the performance of \ours. Our approach outperforms the baselines in most tasks. Details are shown in Appendix \ref{add_baseline}.

\subsection{Ablation Study}

\begin{table}[!htb]
\centering
\resizebox{\columnwidth}{!}{%
\begin{tabular}{lccccc}
\toprule
Method & R52            & Reuters        & AbsQA  & MCQA & Math           \\ \midrule
\multicolumn{6}{c}{Budget Setting 1}                                                                         \\ \midrule
Random FT CoT     & \textbf{59.70} & 27.35          & 52.57          & 76.09          & 15.20          \\
Balance FT CoT    & 51.47          & 27.12          & 52.22          & 75.98          & 16.29          \\
Active FT CoT     & 59.49          & 29.75          & 53.14          & 76.64          & 15.61          \\ \midrule
\ours (N)         & 59.62          & 28.09          & 52.70          & 76.60          & 16.34          \\
\ours (A)         & 58.93          & \textbf{32.95} & \textbf{53.20} & \textbf{77.17} & \textbf{17.42} \\ \midrule
\multicolumn{6}{c}{Budget Setting 2}                                                                         \\ \midrule
Random FT CoT     & 64.88          & \textbf{33.42} & 53.71          & 72.92          & 15.19          \\
Balance FT CoT    & 60.55          & 32.79          & 50.29          & 76.29          & 15.73          \\
Active FT CoT     & 64.54          & 31.33          & 53.05          & 76.26          & 15.91          \\ \midrule
\ours (N)         & 59.35          & 32.76          & \textbf{53.86} & 76.17          & 17.59          \\
\ours (A)         & \textbf{65.84} & 32.77          & 53.49          & \textbf{77.11} & \textbf{17.59} \\ \bottomrule
\end{tabular}
}
\vspace{-0.8em}
\caption{\label{table:abla_results} \textbf{Effects of active learning and adaptive balancing in \ours framework.} Results of fine-tuned student models on five datasets outperform methods with only balancing (Balance FT CoT), with only active learning (Active FT CoT).}
\end{table}

After showing the superiority of our overall framework, our next step is to verify the effectiveness of each component in the proposed method. We compare our framework with the ablated methods: (1) \textbf{Balance Finetune CoT}. We adopt a naive balancing policy to construct the training set and query the teacher model to compose inputs in the tail domains. We randomly sample examples from head domains to make sure they are not over-represented. (2) \textbf{Active Finetune CoT}. We only keep the active learning component but remove the data augmentation part. In details, we calculate the IFD scores for all examples in our original dataset and select the highest IFD scores to satisfy the budget number. Note that this ablation method is equivalent to the SoTA active learning method: Superfiltering \cite{li2024superfiltering}. The experiment setting is similar to the setup in Section \ref{sec:overall_results}, and we present the performance of each method with two budget settings in Table \ref{table:abla_results}.

\noindent \textbf{Both active selection and adaptive balancing bring salient performance boost} \quad
From Table \ref{table:abla_results}, we find that our \ours (A) method obtains the best performance in 7/10 comparison cases, which demonstrates the effectiveness of each framework component. We notice that by simply adding the active learning strategy (Random Finetune CoT vs. Active Finetune CoT), the fine-tuned student model can achieve a performance boost in most cases, with an average relative improvement of 1.43\%. This observation is consistent with the findings in previous work for Bert models \cite{devlin2019bert} on the long-tailed data \cite{dor2020active}.

However, when we add data augmentation from the teacher with the naive balancing policy (Balance Finetune CoT vs. Random Finetune CoT, \ours (N) vs. Active Finetune CoT), this operation does not substantially improve performance. This finding suggests the superiority of our adaptive balancing policy. 

\begin{figure}[!t]
    \centering
    \begin{subfigure}[b]{0.48\columnwidth}
        \centering
        \includegraphics[width=\textwidth]{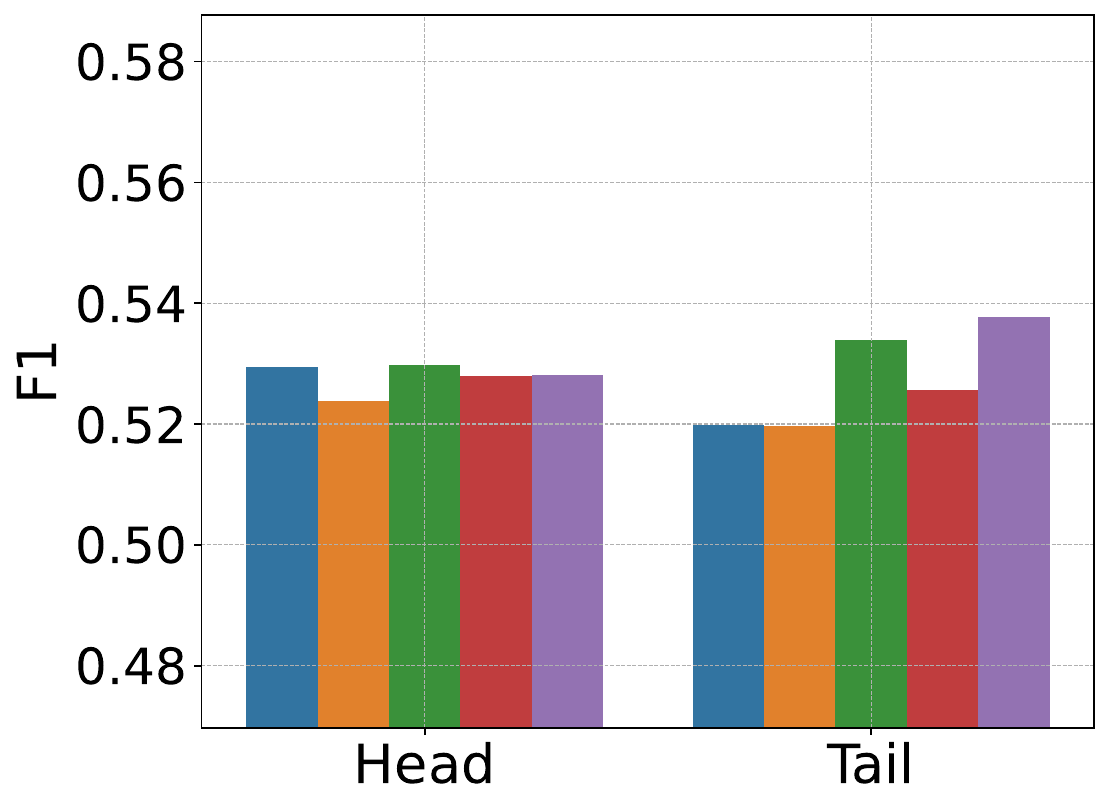}
        \caption{AbstractiveQA}
        \label{fig:abstractive_head}
    \end{subfigure}
    \hfill
    \begin{subfigure}[b]{0.48\columnwidth}
        \centering
        \includegraphics[width=\textwidth]{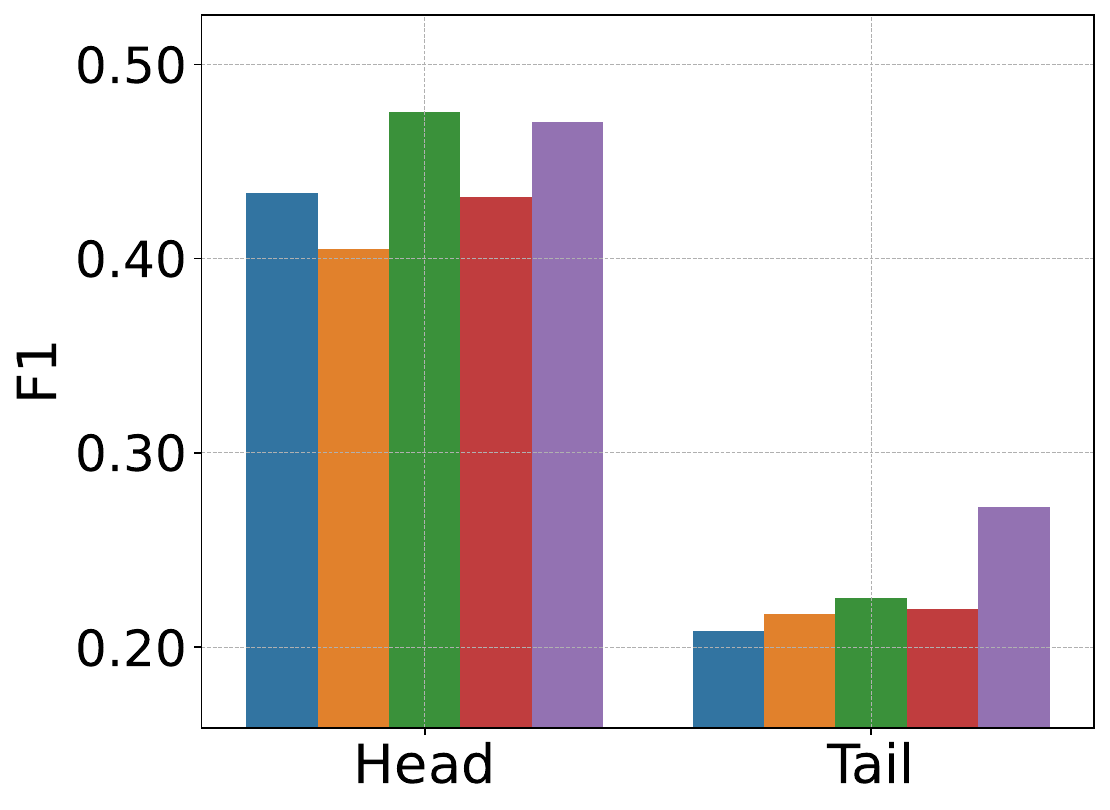}
        \caption{Reuters}
        \label{fig:reuters_head}
    \end{subfigure}
    \vfill
    \begin{subfigure}[b]{0.48\columnwidth}
        \centering
        \includegraphics[width=\textwidth]{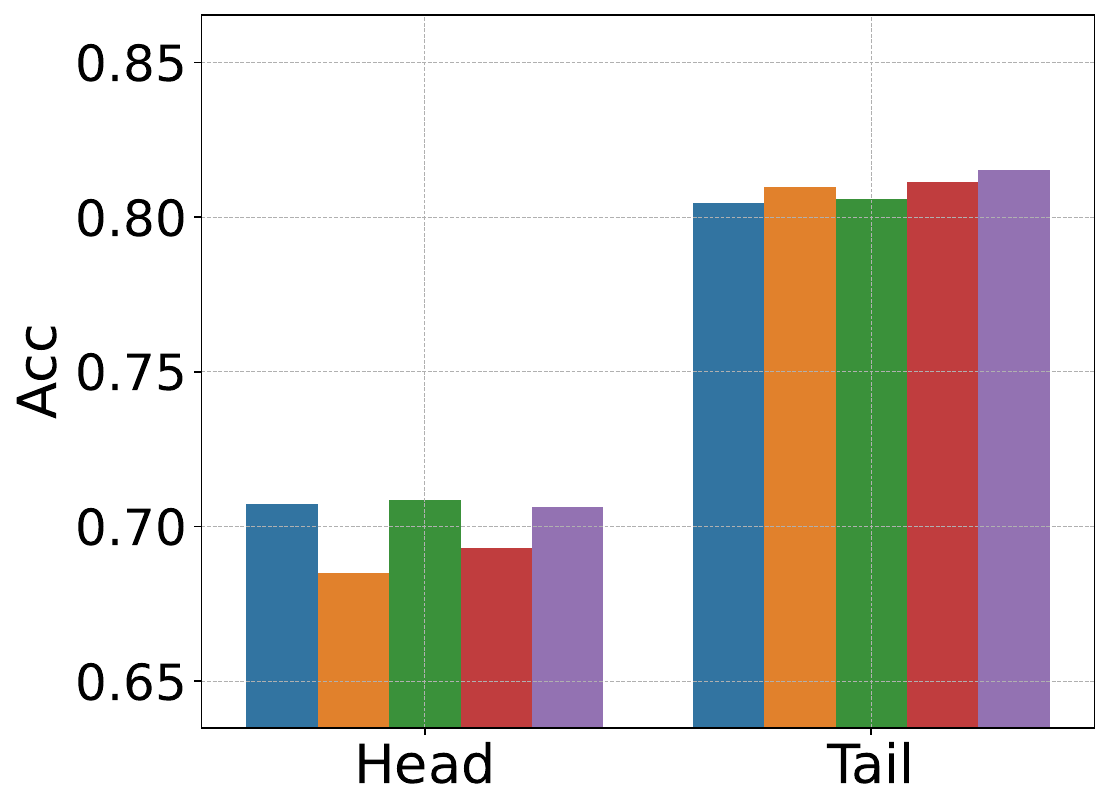}
        \caption{MultichoiceQA}
        \label{fig:multichoiceQA_head}
    \end{subfigure}
    \hfill
    \begin{subfigure}[b]{0.48\columnwidth}
        \centering
        \includegraphics[width=\textwidth]{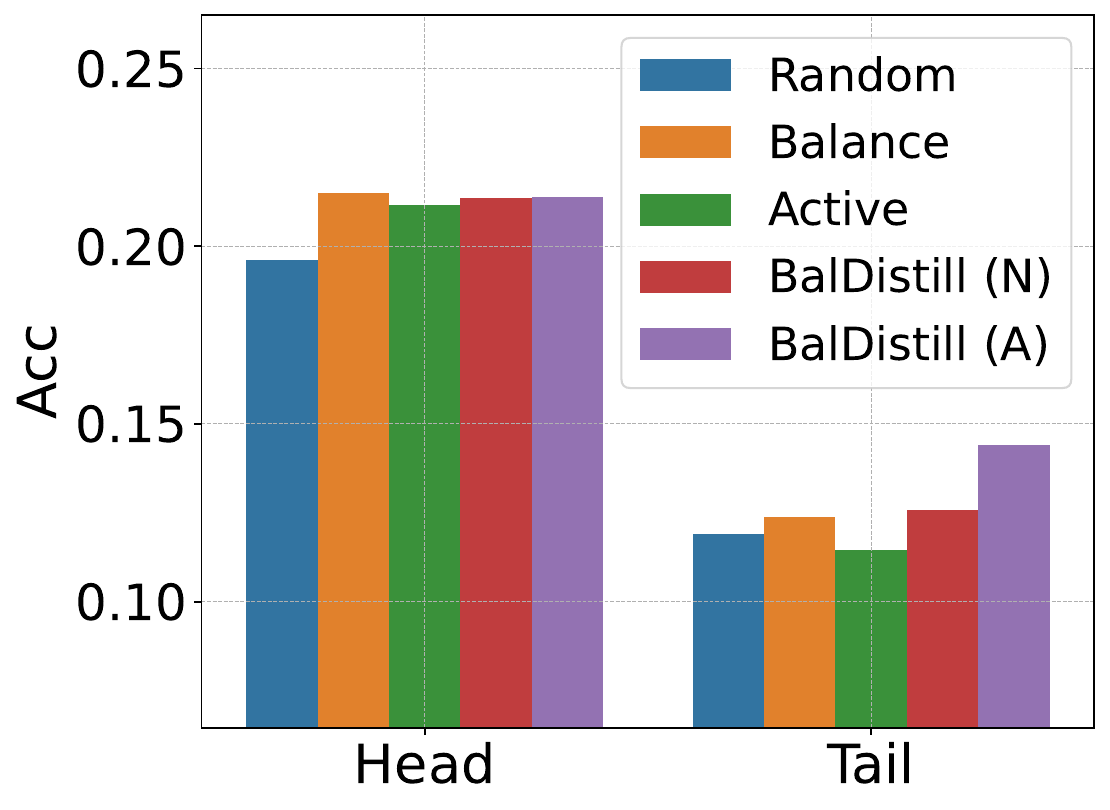}
        \caption{Math}
        \label{fig:math_head}
    \end{subfigure}
    \vspace{-0.5em}
    \caption{\textbf{Performance of proposed method \ours and ablated methods on head and tail domains.} \ours (A) can achieve better results on head domains and outperform the Active FT CoT method on tail domains, which demonstrates the effectiveness of each component in our \ours (A) framework.}
    \label{fig:accuracy_ablation}
\end{figure}

To probe the detailed reasons for the result patterns above, we visualize the macro-average performance of these methods on inputs from head and tail domains in Figure \ref{fig:accuracy_ablation}. The splitting criteria for each dataset can be found in Appendix \ref{data_details}. We find that for methods with naive balancing policy (Balance Finetune CoT and \ours (N)), there exists a significant performance drop on head domains due to filtering a large proportion of data, and our method with adaptive balancing can achieve comparable performance on head domains. The observation suggests the effectiveness of active selection for head domains and the importance of adaptive balancing for the fine-tuned student to select the uncertain ones precisely.

For performance in tail domains, our proposed method with adaptive balancing and teacher augmentation could achieve the best average results, even better than the naive balancing method. We conjecture that since we do not verify the correctness of teacher-generated samples and rationales in tail domains. While teacher-generated samples induce more knowledge, more synthetic data can lead to more inevitable noise. Adaptive balancing achieves the best trade-off between inducing more knowledge and less noise in the tail domains.

\subsection{Generalization Analysis}

The ablation study demonstrates the effectiveness of the active learning and adaptive balancing. Then, we ask whether our proposed method is robust enough to experiment with different hyperparameters, student models, or budget settings.

\subsubsection{Generalizations on Student models}

\begin{table}[!htb]
\centering
\resizebox{\columnwidth}{!}{%
\begin{tabular}{lccccc}
\toprule
Method & R52            & Reuters        & AbsQA  & MCQA & Math           \\ \midrule
\multicolumn{6}{c}{Llama3 Budget Setting 1}                                                                  \\ \midrule
Random FT CoT           & \textbf{59.70} & 27.35          & 52.57          & 76.09          & 15.20          \\
Active FT CoT           & 59.49          & 29.75          & 53.14          & 76.64          & 15.61          \\
\ours (A)         & 58.93          & \textbf{32.95} & \textbf{53.20} & \textbf{77.17} & \textbf{17.42} \\ \midrule
\multicolumn{6}{c}{Llama2 Budget Setting 1}                                                                  \\ \midrule
Random FT CoT           & 49.83          & 23.97          & 46.26          & 58.69          & 3.43           \\
Active FT CoT           & 46.88          & 24.06          & 47.07          & 58.68          & 3.82           \\
\ours (A)         & \textbf{58.33} & \textbf{25.51} & \textbf{47.55} & \textbf{59.14} & \textbf{4.21}  \\ \midrule
\multicolumn{6}{c}{Llama3 Budget Setting 2}                                                                  \\ \midrule
Random FT CoT           & 64.88          & \textbf{33.42} & 53.71          & 72.92          & 15.19          \\
Active FT CoT           & 64.54          & 31.33          & 53.05          & 76.26          & 15.91          \\
\ours (A)         & \textbf{65.84} & 32.77          & \textbf{53.49} & \textbf{77.11} & \textbf{17.59} \\ \midrule
\multicolumn{6}{c}{Llama2 Budget Setting 2}                                                                  \\ \midrule
Random FT CoT           & 56.45          & 23.75          & 48.95          & 58.91          & 3.84           \\
Active FT CoT           & 53.16          & \textbf{27.12} & 48.27          & \textbf{59.20} & 3.52           \\
\ours (A)         & \textbf{58.17} & 27.07          & \textbf{49.45} & 58.64          & \textbf{4.54}  \\ \bottomrule
\end{tabular}
}
\vspace{-0.8em}
\caption{\label{table:diff_size_results} \textbf{Effects of student model scales and budget numbers.} Macro-averages the proposed and baseline method results when considering Llama2 and Llama3 as student models with varying two budget settings.}
\end{table}

We first evaluate whether our method could be generalized to student models with different reasoning abilities or with different budget numbers. In this part, we additionally evaluate our \ours (A) on Llama2-7B models, which have a smaller model size and fewer tokens, in two budget settings (the details of each dataset are in Table \ref{table:dataset}). We present the fine-tuning results of our proposed framework and baseline methods on the Llama2 and Llama3 student models in Table \ref{table:diff_size_results}.

\noindent \textbf{\ours exhibits robust improvement with various budget settings or student models.} \quad From Table \ref{table:diff_size_results}, we observe that fine-tuning with the Llama3-8B student model leads to much better performance than the Llama2-7B model, especially on tasks with complex reasoning (Math, Multi-choice QA), indicating that the student with a larger model size or a better reasoning ability will yield better fine-tuning results. This observation is consistent with previous findings in \citet{ho2022large, hsieh2023distilling}. Our \ours (A) consistently outperforms other baseline methods on both Llama2-7B and Llama3-8B as student models in most cases under two budget settings, which also verifies the generalizability of our \ours (A) on different student models or different budget numbers.

\subsubsection{Sensitivity Analysis}

\begin{figure}[!t]
    \centering
    \begin{subfigure}[b]{0.48\columnwidth}
        \centering
        \includegraphics[width=\textwidth]{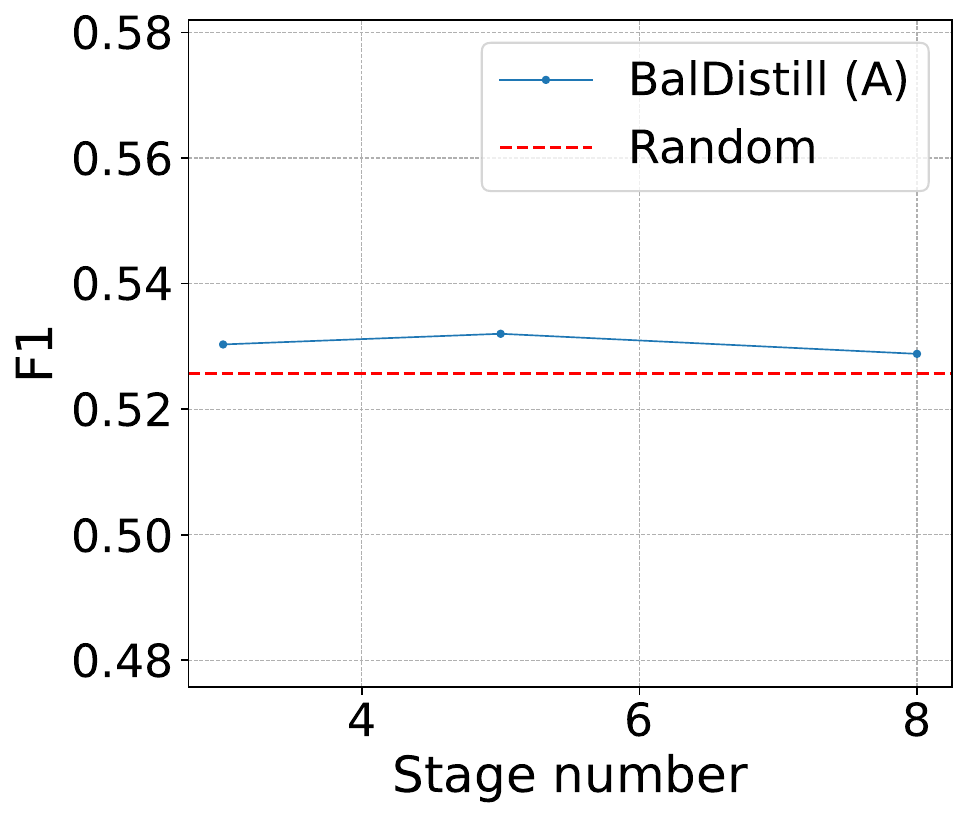}
        \vspace{-1em}
        \caption{AbstractiveQA}
        \label{fig:abs_stage}
    \end{subfigure}
    \hfill
    \begin{subfigure}[b]{0.48\columnwidth}
        \centering
        \includegraphics[width=\textwidth]{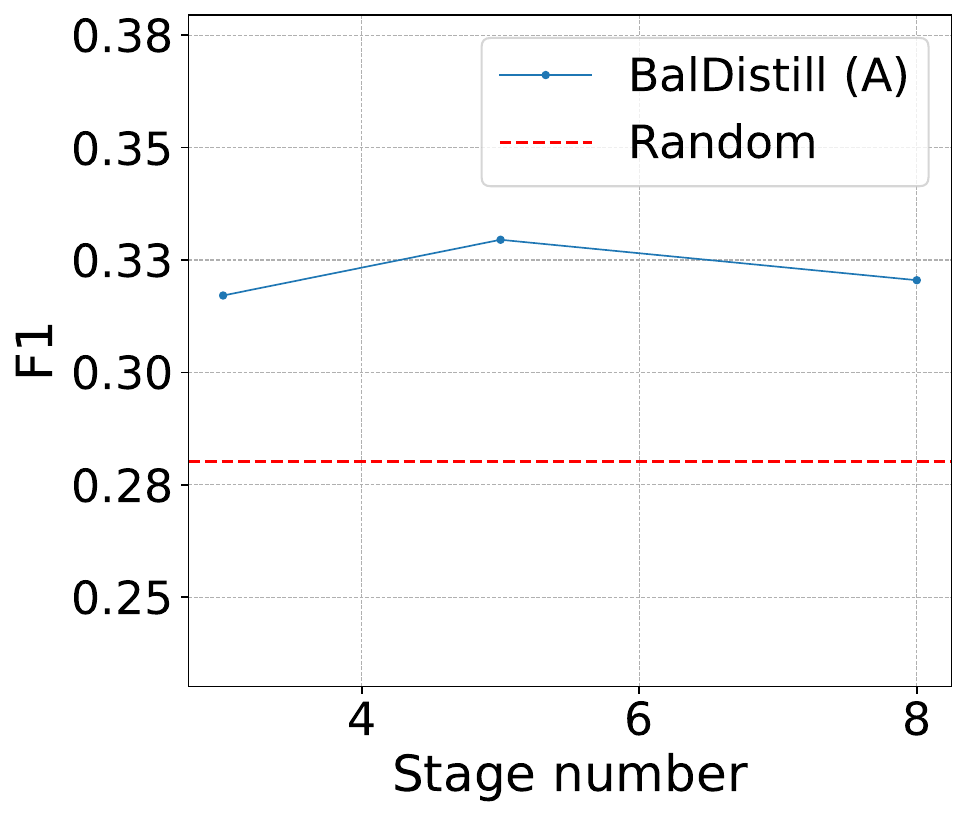}
        \vspace{-1em}
        \caption{Reuters}
        \label{fig:reuters_stage}
    \end{subfigure}
    \vfill
    \begin{subfigure}[b]{0.48\columnwidth}
        \centering
        \includegraphics[width=\textwidth]{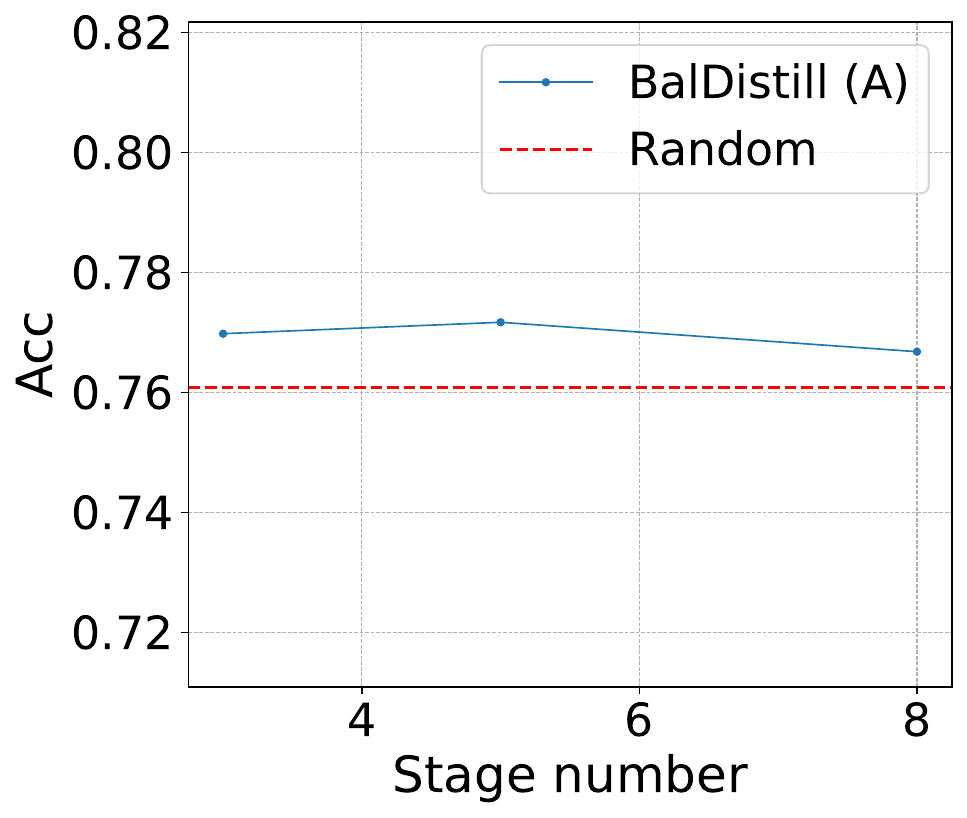}
        \caption{MultiChoiceQA}
        \label{fig:mcqa_stage}
    \end{subfigure}
    \hfill
    \begin{subfigure}[b]{0.48\columnwidth}
        \centering
        \includegraphics[width=\textwidth]{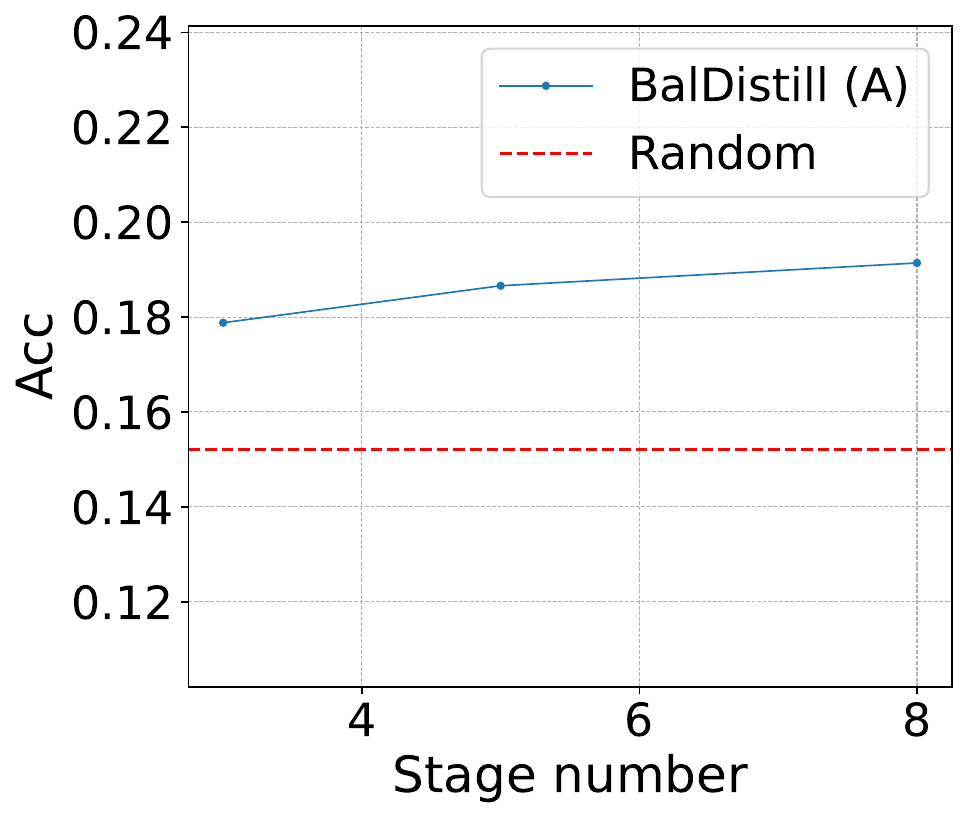}
        \caption{Math}
        \label{fig:math_stage}
    \end{subfigure}
    \vspace{-0.5em}
    \caption{\textbf{Influence of stage number choices on \ours across datasets.} Our proposed method consistently obtains better results than the random fine-tune baseline method with varying stage numbers.}
    \vspace{-1em}
    \label{fig:accuracy_stage}
\end{figure}

We next investigate how the choice of stage number: $K$ will influence the performance of our framework. We experiment with the same setup as in Section \ref{sec:overall_results} but with varying stage numbers among $\{3, 5, 8\}$. We visualize the results (macro-averages) of \ours (A) and the baseline method Random Finetune CoT in Figure \ref{fig:accuracy_stage}

Figure \ref{fig:accuracy_stage} shows that the fine-tuning results of \ours (A) could be affected by the stage number to some extent, but our proposed method can consistently outperform the baseline method with different stage numbers, demonstrating the effectiveness and robustness of \ours (A). 
\section{Conclusions}
  
In this paper, we propose a novel framework \ours to enhance performance on long-tail datasets in the current teacher-student knowledge distillation process. 
Our framework is a multi-stage pipeline, and at each stage, we call the student models to actively select the representative examples from head domains while prompting the teacher to generate synthetic examples for tail domains. 
With a fixed budget restriction for calling the teacher, our extensive empirical evaluations show that our framework can significantly increase fine-tuning results across multiple datasets. Furthermore, we demonstrate the effectiveness of all framework components through ablation studies. 

\section*{Acknowledgments}

Zhou, Wang, Liu and Huang are supported by DARPA Transfer from Imprecise and Abstract Models to Autonomous Technologies (TIAMAT) 80321, National Science Foundation NSF-IIS-2147276 FAI, DOD-ONR-Office of Naval Research under award number N00014-22-1-2335, DOD-AFOSR-Air Force Office of Scientific Research under award number FA9550-23-1-0048, DOD-DARPA-Defense Advanced Research Projects Agency Guaranteeing AI Robustness against Deception (GARD) HR00112020007, Adobe, Capital One and JP Morgan faculty fellowships. Zhu and Koutra are supported by the National Science Foundation CAREER Grant No. IIS 1845491.  Any opinions,
findings, and conclusions or recommendations expressed in this
material are those of the author(s) and do not necessarily reflect the
views of the National Science Foundation or other funding parties.

\section*{Limitations}

In our work, we use the IFD score as the metric for active selection for the student model. In addition to IFD scores, we can try other metrics, such as maximum entropy \cite{settles2009active} or K-means diversity \cite{yuan2020cold}. However, previous work has shown that the IFD score is more effective in selecting data for sequence-level fine-tuning than other metrics \cite{li2024superfiltering, li2024quantity}.

We have verified the effectiveness of our framework on multiple student models and various long-tailed datasets. Other sequence-level KD methods still use more complex loss functions \cite{hsieh2023distilling} or augment the generated rationales \cite{shridhar2023distilling}. Our data manipulation framework complements these KD methods, aiming to achieve more robust results on long-tailed datasets with a fixed budget. Moreover, our method focuses on sequence-level KD for black-box LLMs, so we do not incorporate the KD method for white-box LLMs as a baseline method \cite{gu2023minillm, dai2023long}. We will leave the exploration of combining our framework with more advanced KD methods for the future.

Furthermore, our experiments only focus on the decoder-only student models: Llama3 and Llama2. Incorporating more encoder-decoder models such as FLAN-T5 \cite{flant5} would benefit future studies.

Another future direction for our paper is to explore the application of knowledge distillation in Large Vision-Language Models (LVLMs)~\cite{liu2024visual, liu2024improved, bai2023qwen, sun2023aligning, zhou2024calibrated, wang2024enhancing, lin2024vila, zhu2024multimodal,liu2024c}. In this paper, we have focused on experiments related to knowledge distillation in Large Language Models (LLMs). In future work, we aim to use knowledge distillation to further address the issue of hallucination~\cite{liu2023aligning, cui2023holistic, wang2024mementos} in small LVLMs such as LLaVA-7b~\cite{liu2024visual} and VILA-7b~\cite{lin2024vila}.

\bibliographystyle{acl_natbib}
\bibliography{anthology, custom}

\appendix
\section{Dataset Construction}
\label{data_details}

\noindent \textbf{R52 \& Reuters} We use the original R52 and Reuters dataset. In Figure \ref{table:abla_results}, we treat domains (labels) with more than 50 instances in the training dataset as the head domains and the others as tail domains. 

\noindent \textbf{Multi-choice QA} For Multi-choice QA, we merge 10 multichoice QA datasets together, including Race, OBQA, MCTest, ARC-easy, ARC-hard, CQA, QASC, PIQA, SIQA, Winogrande ~\cite{dai2023long}. For training samples, we downsample the 10 datasets following a Zipf distribution with power value $\alpha = 2.0$ ~\cite{dai2023long}. Since Race has 5$\times$ more training samples than other datasets, we downsample its training and testing set to 1/3 of the samples using random sampling. The detailed statistics of each multichoice qa dataset is shown in Table \ref{tab:category-stats}.We select Race, Winogrande, SIQA and CQA as the head domains and others as tail domains for experiments in Figure \ref{table:abla_results}. 

\noindent \textbf{Abstractive QA} For Abstractive QA, we merge 5 abstractive QA datasets together, including NarQA, NQOpen, Drop, QAConv, TweetQA  ~\cite{dai2023long}. Since the total train set and test set are very large, for efficiency concerns, we randomly sample 10000 samples from them for both train and test sets.  The detailed statistics of each multichoice qa dataset is shown in Table \ref{tab:category-stats}. We select NarQA, NQOpen, and Drop as the head domains and others as tail domains for experiments in Figure \ref{table:abla_results}.

\noindent \textbf{Math} We use the Math dataset from ~\cite{hendrycks2021measuring}, which consists of 7 categories: Algebra, Intermediate Algebra, Prealgebra, Geometry, Number Theory, Counting \& Probability and Precalculus. In order to investigate GPT4's reasoning ability on MATH problems and how much can its reasoning be taughts to the student model, we removed the reasoning procedures in Math dataset and only keep its final answer as the label. Since the original dataset distribution is as follows does not follow long tail distribution, we down-sample the training sets of all categories following a Zipf distribution with power value
$\alpha = 1.1$, similar to ~\cite{dai2023long}. The final distribution of the datsaet is shown in Table \ref{tab:category-stats}. We select Algebra, Intermediate Algebra, and Prealgebra as the head domains and others as the tail domains for experiments in Figure \ref{table:abla_results}.

\begin{figure}[t]

      \subfloat[Reuters]{\includegraphics[width=0.49\textwidth]{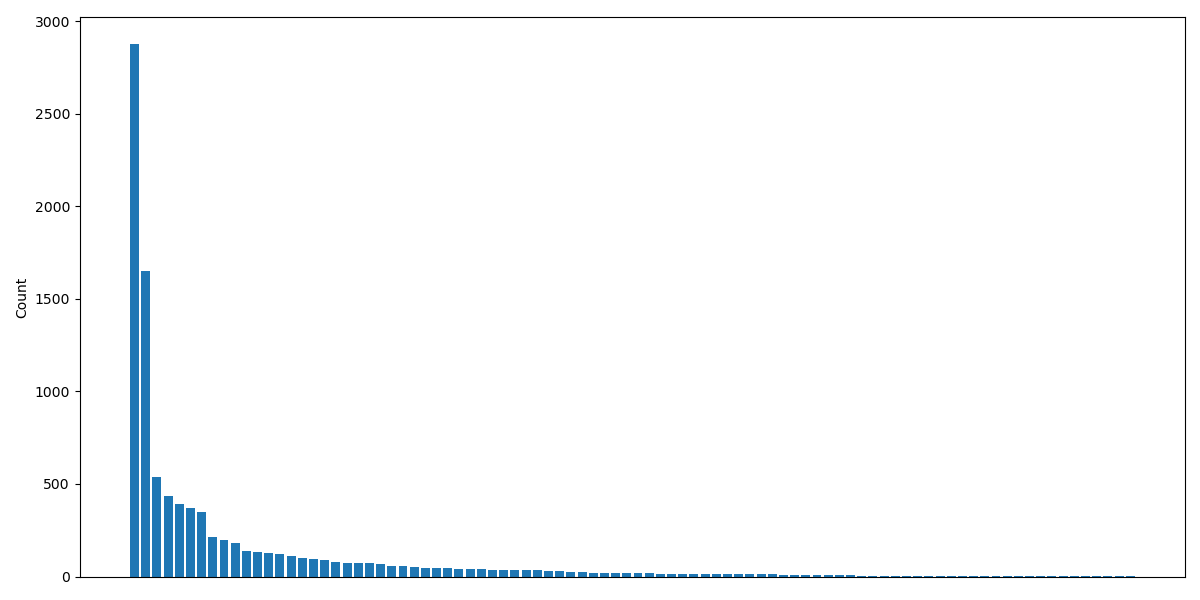}}

    ~
    \subfloat[Math]{\includegraphics[width=0.49\textwidth]{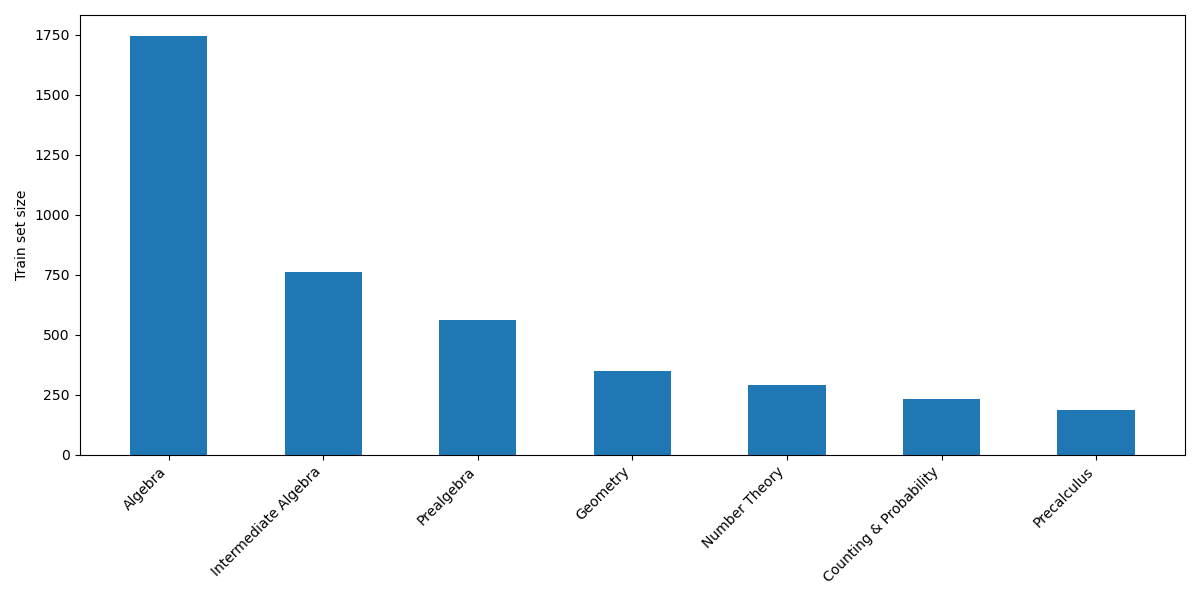}}

	\centering
\caption{Example Dataset Distribution: The datasets we use exhibit long-tail distributions.}
    \label{fig:dataset-long}
\end{figure}

\begin{table}[t!]
\caption{Detailed statistics of each dataset per category. }
\label{tab:category-stats}
\centering
\resizebox{\columnwidth}{!}
{
\begin{tabular}{l@{\hskip5pt}r@{\hskip5pt}r@{\hskip5pt}r}
\toprule
   \textbf{Dataset} & \textbf{Category} & \textbf{Train set size} & \textbf{Test set size}  \\
\midrule 

 \multirow{9}{*}{\textbf{Multi-choice QA}} &   Race & 4735 &  1629\\ 

    & OBQA & 580 & 500 \\

    & MCTest & 342 & 320 \\

    & ARC-easy & 395 & 570 \\

    & ARC-hard & 317 & 299 \\

    & CQA & 1034 & 1221 \\

   &  QASC & 653 & 926 \\

   & SIQA & 2077 & 1954 \\

   & PIQA & 494 & 1838 \\

   & Winogrande & 2634 & 1267 \\ \hline

  \multirow{5}{*}{\textbf{Abstractive QA}} &  NarQA  & 1999 &  2244\\ 
    
  &  NQOpen & 4441 &  3434\\

  &  Drop & 2525 &  2891\\

  &  QAConv& 751 &  1079\\

  &  TweetQA  & 284 &  352\\ \hline

    \multirow{7}{*}{\textbf{Math}} &  Algebra & 1744 &  1187 \\ 

   &  Intermediate Algebra & 763 &  903 \\ 

   & Prealgebra & 561 &  871 \\ 

    & Geometry & 349 &  479 \\ 

  &  Number Theory & 290 &  540 \\ 

  &  Counting \& Probability & 231 &  474 \\ 

  &  Precalculus & 187 &  546 \\

\bottomrule
\end{tabular}

}
\end{table}

\section{Implementation Details}
\label{sec:implementation}

We use greedy search in decoding for all teacher annotations, as in the previous work \cite{ho2022large} and use stochastic temperature sampling with the same temperature value of 0.9 in synthetic data generation in Section \ref{sec:teacher_aug}.

We use the zero-shot prompts for the teacher to give the rationales and the few-shot ICL to generate the synthetic tail samples. The prompts are shown in Tables \ref{table:prompt_syn_1}, \ref{table:prompt_syn_2} and \ref{table:prompt}. We call the gpt-4 function from OpenAI to obtain teacher responses.

For the fine-tuning of the student model, we base our implementation on the Pytorch\footnote{\url{https://pytorch.org/}}, Huggingface transformer\footnote{\url{https://huggingface.co/}}, and the Lora fine-tuning codebase \footnote{\url{https://github.com/georgian-io/LLM-Finetuning-Toolkit/tree/main}}. We use AdamW as our optimizer with a learning rate of $2\mathrm{e}{-4}$ and a weight decay of 0.03 with linear scheduler, batch size of 16, and trained for 8 epochs. For other hyper-parameters, we set rank and dropout in Lora fine-tuning to 8 and 0.1, respectively.

\section{Additional Baseline Results}
\label{add_baseline}

Besides the baselines mentioned in Section \ref{sec:overall_results}, we also include two additional state-of-the-art methods for multitask learning and addressing class imbalance challenges. The first method is Glee \cite{zhang2022making}, which leverages prompt tuning on masked tokens to handle long-tailed classification tasks. We adapted the classification head of Glee to make it applicable for generation tasks. However, since the final answers for AbstractiveQA consist of multiple words with variable lengths, Glee cannot be leveraged for AbstractiveQA.
The second method is MFTCoder \cite{liu2024mftcoder}, which proposes various loss functions to address challenges in multitask learning. For our experiment, we treat each domain as a separate task within their framework. Note that we employ instruction tuning for both MFTCoder and \ours, while using prompt tuning for Glee.

Below we present the macro metrics of the baselines and \ours on each dataset. Llama2 is used as backbones for all of the methods. 

\begin{table}[h!]
\resizebox{\columnwidth}{!}{%
\centering
\begin{tabular}{lccccc}
\hline
\textbf{Dataset} & \textbf{R52} & \textbf{Reuters} & \textbf{AbsQA} & \textbf{MCQA} & \textbf{Math} \\ \hline
MFTCoder          & 10.49        & 6.63            & 33.76          & 58.62         & 3.22          \\ \hline
Glee              & 46.87        & 23.96           & N/A            & 57.88         & \textbf{4.37}          \\ \hline
BalDistill (A)    & \textbf{58.33}        & \textbf{25.51}           & \textbf{47.55}          & \textbf{59.14}         & 4.21          \\ \hline
\end{tabular}
}
\caption{\label{tab:result_baseline_add} Performance of \ours and two additional baselines across 5 datasets.}
\end{table}

From the results in Table \ref{tab:result_baseline_add}, it is evident that our approach outperforms the MFTcoder and Glee on most tasks. MFTCoder's underperformance can be attributed to its use of validation loss gradients (as described in Equation 4 of their paper) to adjust training loss, which leads to unstable learning. 
In our experimental setup, particularly for tail labels in the R52 and Reuters datasets, we often had only one or two examples in the validation set. This scarcity can cause large gradients in the validation data, potentially leading to loss explosion during fine-tuning on these datasets. Glee's limitations results in its failure to utilize information from teacher rationales. In contrast, our method leverages data augmentation from teachers to elicit more knowledge and enhance fine-tuning for tail domains. This approach allows us to better capture and utilize the expertise embedded in the teacher models, resulting in improved performance across various tasks.

\begin{table*}[ht]
    \begin{subtable}{1\textwidth}
        \centering
        \begin{tabular}{m{40em}}
        \toprule
        You are provided with a dataset named R52, which is specifically designed for text classification tasks. The objective is to accurately predict the topic of news stories from a predefined list of topics. The topic of this dataset includes: copper, livestock, gold, money-fx, tea, ipi, trade, cocoa, iron-steel, reserves, zinc, nickel, ship, cotton, platinum, alum, strategic-metal, instal-debt, lead, housing, gnp, sugar, rubber, dlr, tin, interest, income, crude, coffee, jobs, meal-feed, lei, lumber, gas, nat-gas, veg-oil, orange, heat, wpi, cpi, earn, jet, potato, bop, money-supply, carcass, acq, pet-chem, grain, fuel, retail, cpu. Please write a short news story with the topic \{domain\} and give the step-by-step rationale. This should be a self-contained story, mirroring the style and content of real-world news articles. Here are some examples with the topic \{domain\}: \\
        \{demonstrations\}\\
        Please compose a news story with the topic \{domain\} with a similar format as the example. Paraphrase your title before outputting it. Your news story should be brief and contained within one paragraph:
        \\
        \bottomrule
        \end{tabular}
        \caption{R52}
    \end{subtable}

        \begin{subtable}{1\textwidth}
        \centering
        \begin{tabular}{m{40em}}
        \toprule
        You are provided with a dataset named reuters, which is specifically designed for text classification tasks. The objective is to accurately predict the topic of news stories from a predefined list of topics. The topic of this dataset includes: acq, rubber, lead, money-supply, income, l-cattle, crude, cpu, palmkernel, jobs, money-fx, instal-debt, rand, castor-oil, coffee, strategic-metal, nat-gas, oat, tea, corn, yen, soy-oil, grain, groundnut-oil, gas, cpi, cocoa, nzdlr, soybean, rapeseed, retail, sun-meal, coconut, jet, copper, sorghum, carcass, heat, hog, ipi, potato, lin-oil, oilseed, alum, gnp, meal-feed, fuel, barley, ship, rape-oil, cotton-oil, sunseed, palm-oil, soy-meal, naphtha, nkr, trade, palladium, lei, wheat, bop, interest, earn, reserves, housing, veg-oil, groundnut, tin, dlr, gold, copra-cake, wpi, livestock, zinc, sugar, rye, pet-chem, dmk, dfl, orange, iron-steel, nickel, sun-oil, lumber, rice, propane, platinum, silver, cotton, coconut-oil. Please write a short news story with the topic \{domain\} and give the step-by-step rationale. This should be a self-contained story, mirroring the style and content of real-world news articles. Here are some examples with the topic \{domain\}:\\
        \{demonstrations\} \\
        Please compose a news story with the topic \{domain\} with a similar format as the example and your news story should be brief and contained within one paragraph:
        \\
        \bottomrule
        \end{tabular}
        \caption{Reuters}
    \end{subtable}

    \caption{\label{table:prompt_syn_1} Prompts of generating synthetic data for tail domains from the teacher for R52 and reuters datasets. }
\end{table*}

\begin{table*}[ht]
    \begin{subtable}{1\textwidth}
    \centering
    \begin{tabular}{m{40em}}
        \toprule
        You are provided with a multiple-choice question and answering dataset composed by various QA datasets. The objective is to accurately select one from the given choices according to the question content. Please compose a question as well as the corresponding choices and answers as the examples from a QA dataset: \{domain\}. This should be a question, mirroring the style and content of examples with the true real-world knowledge. Here are some examples from the QA dataset: \{domain\}:\\
        \{demonstrations\} \\
        Please compose a question for the dataset: \{domain\} with a similar format as the example. It means if the example contains the in-context "passage", you should also write an in-context "passage" with the question information. Your question and choices should be brief and contained within one paragraph:
        \\
        \bottomrule
        \end{tabular}
    \caption{Multi-choice QA}
    \end{subtable}

    \begin{subtable}{1\textwidth}
    \centering
    \begin{tabular}{m{40em}}
        \toprule
        You are provided with an abstractive question answering dataset composed by various QA datasets. The objective is to accurately generate an answer according to the question content. Please compose a question and the corresponding answer as the examples from a QA dataset: \{domain\}. This should be a question and answer, mirroring the style and content of examples with the true real-world knowledge. Here are some examples from the QA dataset: \{domain\}: \\
        \{demonstrations\} \\
        Please compose a question and the corresponding answer for the dataset: \{domain\} with a similar format as the example. It means if the example contains the in-context "passage", you should also write an in-context "passage" with the question information. Please note that the answer should only contain a few words. Your question and answer should be brief and contained within one paragraph:
        \\
        \bottomrule
        \end{tabular}
    \caption{Abstractive QA}
    \end{subtable}
    
    \begin{subtable}{1\textwidth}
    \centering
    \begin{tabular}{m{40em}}
        \toprule
        You are provided with a math problem dataset with questions from various math domains. The objective is to accurately generate an answer according to the question content. Please compose a question and the corresponding answer as the examples from a math domain: \{domain\}. This should be a math question and answer, mirroring the style and content of examples with the true real-world knowledge. Here are some examples from the math domain: \{domain\}: \\
        \{demonstrations\} \\
        Please compose a math question and the corresponding answer for the domain: \{domain\}, with a similar format as the example. Please output your final digital answer (no unit) for the question with the format: "the answer is: <answer>". Your question and answer should be brief and contained within one paragraph:
        \\
        \bottomrule
        \end{tabular}
    \caption{Math}
    \end{subtable}
    \caption{\label{table:prompt_syn_2} Prompts of generating synthetic data for tail domains from the teacher for Multi-choice QA, Abstractive QA and Math datasets. }
\end{table*}

\begin{table*}[ht]
    \begin{subtable}{1\textwidth}
        \centering
        \begin{tabular}{m{40em}}
        \toprule
        Below is a news story from the R52 dataset. Please assign a topic to this news story. You must select the topic from this set: copper, livestock, gold, money-fx, tea, ipi, trade, cocoa, iron-steel, reserves, zinc, nickel, ship, cotton, platinum, alum, strategic-metal, instal-debt, lead, housing, gnp, sugar, rubber, dlr, tin, interest, income, crude, coffee, jobs, meal-feed, lei, lumber, gas, nat-gas, veg-oil, orange, heat, wpi, cpi, earn, jet, potato, bop, money-supply, carcass, acq, pet-chem, grain, fuel, retail, cpu. News story: \{input\}. \\
        Take a step-by-step approach in your response, cite sources and give reasoning. Your answer should be brief and contained within one paragraph.
        \\
        \bottomrule
        \end{tabular}
        \caption{R52}
    \end{subtable}

        \begin{subtable}{1\textwidth}
        \centering
        \begin{tabular}{m{40em}}
        \toprule
        Below is a news story from the reuters dataset. Please assign a topic to this news story. You must select the topic from this set: acq, rubber, lead, money-supply, income, l-cattle, crude, cpu, palmkernel, jobs, money-fx, instal-debt, rand, castor-oil, coffee, strategic-metal, nat-gas, oat, tea, corn, yen, soy-oil, grain, groundnut-oil, gas, cpi, cocoa, nzdlr, soybean, rapeseed, retail, sun-meal, coconut, jet, copper, sorghum, carcass, heat, hog, ipi, potato, lin-oil, oilseed, alum, gnp, meal-feed, fuel, barley, ship, rape-oil, cotton-oil, sunseed, palm-oil, soy-meal, naphtha, nkr, trade, palladium, lei, wheat, bop, interest, earn, reserves, housing, veg-oil, groundnut, tin, dlr, gold, copra-cake, wpi, livestock, zinc, sugar, rye, pet-chem, dmk, dfl, orange, iron-steel, nickel, sun-oil, lumber, rice, propane, platinum, silver, cotton, coconut-oil. News story: \{input\}. \\
        Take a step-by-step approach in your response, cite sources and give reasoning. Your answer should be brief and contained within one paragraph.
        \\
        \bottomrule
        \end{tabular}
        \caption{Reuters}
    \end{subtable}

    \begin{subtable}{1\textwidth}
    \centering
    \begin{tabular}{m{40em}}
        \toprule
        Please answer this multiple-choice question by choosing one of the given choices. If you are given a passage, please answer the question according to the passage content. If the passage is not given, please answer the question directly from your knowledge.
        Question: \{input\} \\
        If there is no enough information, you should choose a most possible choice. Take a step-by-step approach in your response, cite sources and give reasoning before sharing final answer in the format: The answer is <selected choice>.
        \\
        \bottomrule
        \end{tabular}
    \caption{Multi-choice QA}
    \end{subtable}

    \begin{subtable}{1\textwidth}
    \centering
    \begin{tabular}{m{40em}}
        \toprule
        Here are a question and the corresponding answer for an abstractive question answering task. Please concisely clarify the rationale behind the answer for this question. If you are given a passage, please use the passage content to clarify the answer. If the passage is not given, please use your knowledge to tell why the answer is reasonable.
        Question: \{input\}. Answer \{label\}. \\
        Take a step-by-step approach in your response and give reasoning. You output should be concise and in one paragraph.
        \\
        \bottomrule
        \end{tabular}
    \caption{Abstractive QA}
    \end{subtable}
    
    \begin{subtable}{1\textwidth}
    \centering
    \begin{tabular}{m{40em}}
        \toprule
        Here are a math question and the ground truth answer for a math problem task. 
        Please concisely clarify the rationale behind the answer for this math question.
        Question: \{input\}. Answer: \{label\}
        \\
        Take a step-by-step approach in your response and give reasoning. 
        You output should be concise and in one paragraph.
        \\
        \bottomrule
        \end{tabular}
    \caption{Math}
    \end{subtable}
    \caption{\label{table:prompt} Teacher annotation prompts for all datasets. Note that for Math, Abstractive QA which has complex reasoning trajectories, we directly provide the ground-truth answer and ask the teacher model to explain the rationale behind the answer and for other datasets, we ask the teacher model to annotate the label and provide the reasoning steps.}
\end{table*}

\begin{table*}[ht]
    \begin{subtable}{1\textwidth}
    \centering
        \begin{tabular}{m{40em}}
            \toprule
            \textbf{Input:} ENERGY GIANT \&lt;EGC> EXPANDS NATURAL GAS OPERATIONS. Energy Giant Corp (EGC) announced plans to expand its natural gas operations in the Midwest, following approval from the Federal Energy Regulatory Commission. The expansion will see EGC increase its natural gas production by 15\%, aiming to meet the growing demand in the region. The company stated that the expansion will be funded by a mix of internal accruals and debt, with the project expected to be operational by the end of the next fiscal year. EGC also highlighted that the expansion aligns with its commitment to sustainable energy solutions, as natural gas is a cleaner-burning fuel compared to coal and oil. The company's shares rose 2\% in early trading following the announcement. \\ \hline
            \textbf{Rationale:} The news story is about Energy Giant Corp (EGC) announcing plans to expand its natural gas operations in the Midwest. The company plans to increase its natural gas production by 15\% to meet the growing demand in the region. The expansion will be funded by a mix of internal accrivals and debt. The company's shares rose 2\% in early trading following the announcement.The story does not mention anything about copper, livestock, gold, money-fx, tea, ipi, trade, cocoa, iron-steel, reserves, zinc, nickel, ship, cotton, platinum, alum, strategic-metal, instal-debt, lead, housing, gnp, sugar, rubber, dlr, tin, interest, income, crude, coffee, jobs, meal-feed, lei, lumber, gas, veg-oil, orange, heat, wpi, cpi, earn, jet, potato, bop, money-supply, carcass, acq, pet-chem, grain, fuel, retail, cpu. However, the story is about the expansion of natural gas operations, which falls under the topic of 'nat-gas'. The answer is nat-gas. \\ \bottomrule
        \end{tabular}
        \caption{The example of teacher synthetic data in R52 dataset for domain: nat-gas.}
    \end{subtable}
    \bigskip
    \begin{subtable}{1\textwidth}
    \centering
        \begin{tabular}{m{40em}}
            \toprule
            \textbf{Input:} A rectangle has a length of 14 units and a width of 8 units. A diagonal is drawn from one corner of the rectangle to the opposite corner. What is the length of the diagonal in units? \\ \hline
            \textbf{Rationale:} The length of the diagonal of a rectangle can be found using the Pythagorean theorem, which relates the lengths of the sides of a right triangle. The theorem states that the square of the length of the hypotenuse (the side opposite the right angle, which is the diagonal in this case) is equal to the sum of the squares of the lengths of the other two sides (the length and width of the rectangle). For this rectangle, the length (l) is 14 units and the width (w) is 8 units. According to the Pythagorean theorem, the length of the diagonal (d) can be calculated as follows: $d^2 = l^2 + w^2$. Plugging in the values gives $d^2 = 14^2 + 8^2 = 196 + 64 = 260$. Taking the square root of both sides to solve for d gives d = $\sqrt{260}$. Factoring out the square root of 4 (which is a perfect square) from under the radical to simplify, we get $d = 2\sqrt{65}$. Thus, the length of the diagonal is $2\sqrt{65}$ units. The answer is: $2\sqrt{65}$. \\ \bottomrule
        \end{tabular}
        \caption{The example of teacher synthetic data in math dataset for domain: geometry.}
    \end{subtable}
    \caption{\label{table:synthetic_examples} Examples of the synthetic input and rationales from the teacher model for the tail domains.}
\end{table*}

\end{document}